\documentclass[preprint,12pt]{elsarticle}
\usepackage{booktabs}
\usepackage[normalem]{ulem}

\usepackage{amssymb}
\usepackage[dvipsnames]{xcolor}
\usepackage{adjustbox}
\usepackage[percent]{overpic}
\usepackage{amsmath}
\usepackage[utf8]{inputenc}
\usepackage{amsmath, amssymb, amsfonts}
\usepackage{longtable}
\usepackage{pdflscape}
\usepackage{bm}
\usepackage{booktabs}
\usepackage{enumitem}
\usepackage{subcaption}
\usepackage{booktabs}
\usepackage{graphicx}
\usepackage{adjustbox}
\usepackage{float}
\usepackage{xcolor}
\usepackage{comment}
\usepackage[table]{xcolor}
\definecolor{lightgray}{gray}{0.93}

\journal{}

\begin{document}

\begin{frontmatter}



\title{FedCC: A Low-Resource Federated Adaptation of Foundation Models for Robust Corpus Callosum localization in Fetal Ultrasound Images} 


\author[1]{Alessandro Di Matteo\corref{cor1}}
\ead{alessandro.dimatteo@unich.it}

\author[1]{Sara Moccia}
\author[3]{Giuseppe Rizzo}
\author[4]{Gianpaolo Grisolia}
\author[5]{Ricciarda Raffaelli}
\author[6]{Lorenzo Vasciaveo}
\author[2]{Francesco d'Antonio\corref{cor2}}

\author[1]{Maria Chiara Fiorentino\corref{cor2}}
\ead{mariachiara.fiorentino@unich.it}

\cortext[cor1]{Corresponding author}

\cortext[cor2]{Co-last}

\affiliation[1]{organization={Department of Innovative Technologies in Medicine \& Dentistry, Università degli Studi "G. D'Annunzio" Chieti - Pescara},
            city={Chieti},
            country={Italy}}
\affiliation[2]{organization={Centre for Fetal care and High Risk Pregnancy, Università degli Studi "G. D'Annunzio" Chieti - Pescara},
            city={Chieti},
            country={Italy}}
\affiliation[3]{organization={Department of Obstetrics and Gynaecology, Sapienza Università di Roma},
            city={Roma},
            country={Italy}}
\affiliation[4]{organization={Department of Obstetrics and Gynaecology},
            city={Mantova},
            country={Italy}}
\affiliation[5]{organization={Department of Obstetrics and Gynaecology, Università di Verona},
            city={Verona},
            country={Italy}}
\affiliation[6]{organization={Department of Obstetrics and Gynaecology, Università di Foggia},
            city={Foggia},
            country={Italy}}

\begin{abstract}

\textbf{Background:} Accurate localization of the corpus callosum (CC) in fetal ultrasound (US) images is crucial for the early identification of neurodevelopmental abnormalities. However, this task remains highly challenging due to the intrinsic limitations of US imaging, including low contrast, speckle noise, and the considerable anatomical variability of the CC. In addition, substantial inter-site variability caused by heterogeneous acquisition protocols and imaging devices further complicates automated analysis. Despite its clinical importance, automatic CC localization in fetal US has received limited attention in the literature, mainly because of these technical challenges and the scarcity of publicly available annotated datasets.

\textbf{Methods:} We propose \textit{FedCC}, a federated learning (FL)-based framework for CC localization in fetal US images, specifically designed for realistic multi-center and resource-constrained clinical settings without requiring data sharing. The framework integrates a frozen DINOv2 backbone with a lightweight YOLO-based detection head. To enable parameter-efficient adaptation, Low-Rank Adaptation (LoRA) modules are incorporated, allowing only a small subset of parameters to be optimized and exchanged among clients. This strategy substantially reduces both computational and communication overhead, making the framework suitable for low-resource environments. The proposed approach was evaluated on a multi-center dataset comprising 10,970 ultrasound frames acquired from 58 pregnant women during routine neurosonographic examinations across three clinical sites using heterogeneous imaging devices.

\textbf{Results \& Discussions:} The proposed framework achieved strong performance in the federated setting. In particular, the combination of DINOv2 and LoRA under the FedAvg strategy achieved an average mAP@50 of 0.857 and an F1-score of 0.803, outperforming both full fine-tuning and encoder-freezing baselines. Notably, the proposed approach reduced the number of trainable parameters to 2.9M compared with 24.4M in full fine-tuning, corresponding to an approximately 8.5$\times$ reduction in communication cost. Furthermore, the federated framework improved generalization across heterogeneous domains, especially on the most challenging client. These findings represent a promising step toward scalable, privacy-preserving, and clinically deployable AI systems for fetal neurosonography, although further validation on larger and more diverse cohorts is still required.

\end{abstract}

\begin{graphicalabstract}
\begin{center}
\vspace{1cm}
\includegraphics[width=1\textwidth]{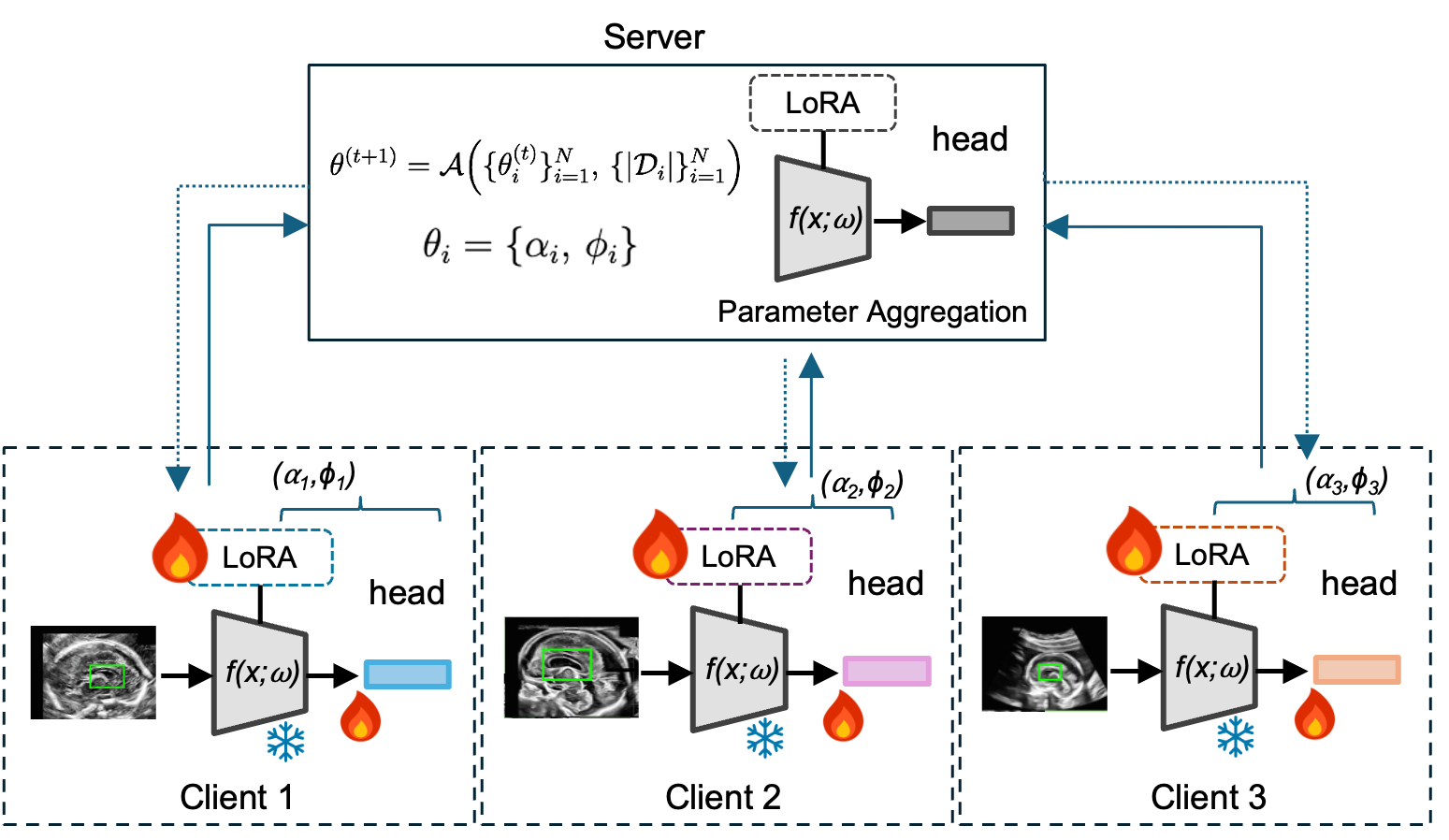}
\end{center}
\end{graphicalabstract}

\begin{highlights}
\item A federated learning framework for corpus callosum detection in fetal ultrasound images is proposed.
\item A frozen DINOv2 backbone is combined with LoRA adapters and a YOLO-based detection architecture.
\item Only LoRA adapters and detection-head parameters are shared across federated clients.
\item Communication costs are reduced by up to 8.5$\times$ compared with full model fine-tuning.
\item An average mAP@50 of 0.857 and an F1-score of 0.803 are achieved on a multi-center fetal ultrasound dataset.

\end{highlights}


\end{frontmatter}



\section{Introduction}

\begin{figure}[H]
    \centering
    \begin{subfigure}[b]{0.48\linewidth}
        \centering
        \includegraphics[width=\linewidth]{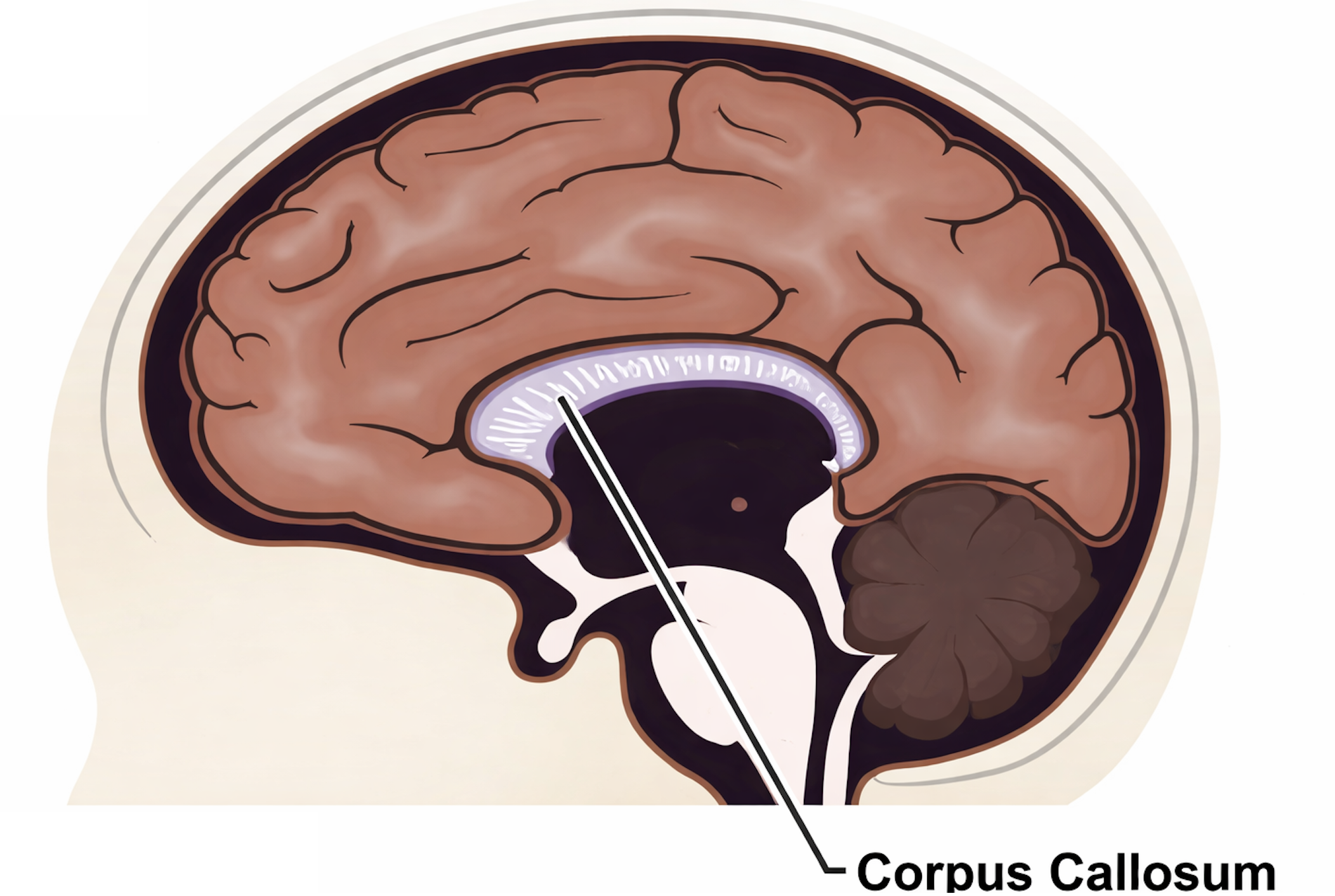}
        \caption{Sagittal view of the fetal brain highlighting the corpus callosum (yellow structure) and surrounding midline structures.}
        \label{fig:cc_anatomy}
    \end{subfigure}
    \hfill
    \begin{subfigure}[b]{0.48\linewidth}
        \centering
        \includegraphics[width=\linewidth]{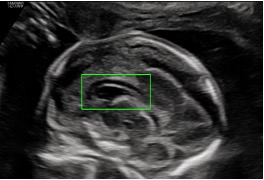}
        \caption{Representative fetal ultrasound frame with the ground-truth bounding box (green) delineating the corpus callosum in the mid-sagittal plane.}
        \label{fig:cc_bbox}
    \end{subfigure}    
    \caption{Anatomical context and localisation target for corpus callosum detection.}
    \label{fig:cc}
\end{figure}

Fetal ultrasound (US) is the standard imaging modality for monitoring fetal development and assessing the anatomy of the central nervous system throughout pregnancy \cite{zhang2020value,paladini2007sonographic}.  During the second and third trimester, neurosonographic examination plays a fundamental role in the early detection of brain abnormalities, enabling timely prenatal counseling and clinical management \cite{wang2024assessment}. Among the anatomical structures evaluated in this context, the corpus callosum (CC) holds particular clinical relevance. As the main white matter commissure connecting the two cerebral hemispheres, the CC is essential for interhemispheric communication and normal neurodevelopment~\cite{lanzarone2025fetal}, and its prenatal assessment is routinely performed on the mid-sagittal plane, where it appears as a thin, curved hypoechoic band above the cavum septi pellucidi (Fig. \ref{fig:cc}).

Accurate localization of this structure is clinically critical, as abnormalities of the CC, including agenesis, dysgenesis, and hypoplasia, represent some of the most common congenital malformations of the fetal central nervous system, frequently associated with neurodevelopmental impairment, chromosomal abnormalities, and additional cerebral anomalies \cite{pilu2016fetal, marathu2024fetal}. Early and reliable detection is therefore essential to guide further diagnostic investigations, genetic counseling, and prognosis estimation \cite{santo2012counseling}.

Despite its clinical importance, robust localization of the CC in fetal US remains challenging. The structure is inherently small, and its morphological appearance changes substantially during gestation, particularly between 18 and 32 weeks, as the CC progressively elongates and thickens \cite{egana2015neurosonographic}. These developmental changes are compounded by the intrinsic limitations of fetal US imaging, including speckle noise, low contrast, acoustic shadowing, and considerable variability introduced by fetal position, maternal body habitus, and probe orientation \cite{meng2020automatic, fiorentino2023review}. In this context, automatic methods for CC localization could meaningfully support clinicians by improving the consistency and efficiency of fetal neurosonographic examinations.
Deep learning (DL)-based approaches have shown considerable promise in automating complex image analysis tasks in fetal US, potentially supporting clinicians by improving consistency, reproducibility, and efficiency across different clinical settings. Several  methods have been proposed for fetal US analysis, including standard plane detection, biometric estimation, and landmark localization ~\cite{fiorentino2023review,torres2022review}. However, realizing their full potential requires training on large and diverse datasets that capture the variability of real-world acquisitions across different centers, scanners, and populations, which is a requirement that may directly conflict with the regulatory and ethical constraints that typically limit direct data sharing between institutions. As a consequence, most existing methods for fetal US image analysis have been developed using data from a single center and a specific US system \cite{burgos2020evaluation, wang2023method}, making them inherently prone to overfitting to a particular acquisition setting and poorly generalizable to external clinical environments.

To address these limitations, recent studies have begun to explore the use of foundation models (FMs), which are pretrained on large and heterogeneous datasets and are therefore able to learn more general and transferable representations \cite{jiao2024usfm, khan2025comprehensive}. Compared with conventional models trained from scratch on limited fetal US datasets, FMs have the potential to better handle the variability introduced by different scanners, acquisition protocols, and clinical centers \cite{meyer2025ultrasam,ma2025tinyusfm}. 
However, adapting foundation models (FMs) to task-specific applications typically relies on fine-tuning strategies that require updating a large number of parameters, resulting in substantial computational and memory demands, particularly in resource-constrained environments. Parameter-Efficient Fine-Tuning (PEFT) approaches address this limitation by introducing only a small set of trainable parameters while keeping most of the pretrained backbone frozen. In this way, PEFT significantly reduces computational and communication costs while preserving the rich and generalizable representations learned during pretraining~\cite{zhang2025adapting,fiorentino2025adapt}.

These characteristics become particularly advantageous in federated learning (FL) settings, where multiple institutions collaboratively train a shared model without exchanging sensitive patient data, thereby preserving privacy and complying with data-sharing regulations~\cite{sheller2020federated}. When integrated with PEFT, only the lightweight adapter parameters need to be exchanged among participating institutions during the federated optimization process, while the FM backbone remains locally stored and frozen. This strategy substantially decreases both the number of trainable parameters and the communication overhead across sites, making the deployment and adaptation of large FMs more practical in distributed clinical environments~\cite{bian2025survey,sheller2020federated}.

Moreover, reducing the number of trainable and transmitted parameters also lowers the computational and energy demands associated with model adaptation, potentially decreasing the environmental impact of AI development and deployment. This perspective is aligned with the European Commission’s guidelines for trustworthy AI, which identify societal and environmental well-being as a key requirement and encourage the design of AI systems that consider their environmental footprint throughout their lifecycle~\cite{eu2024aiact}.
By requiring only lightweight parameter updates, the proposed framework lowers both the computational and communication burden, thereby facilitating participation in collaborative training initiatives even in resource-constrained settings \cite{sheller2020federated,sendra2023generalisability}. Taken together, the integration of FL, FMs, and PEFT represents a principled and promising strategy for developing robust, generalizable systems for automatic CC localization across heterogeneous fetal US datasets.
The main contributions of this work are as follows:

\begin{itemize}
    \item \textit{FedCC}, a FL framework for automatic CC localization that combines a DINOv2-based FM with a YOLO-based detection head. The FM backbone is kept frozen, and only lightweight Low-Rank Adaptation (LoRA) adapter modules \cite{hu2022lora} are trained and shared across participating centers during 
    federation, drastically reducing both the number of trainable parameters and the communication overhead. This design makes the adaptation of large FMs practically feasible in distributed clinical settings while preserving patient privacy.

   \item A systematic evaluation of robustness under multi-device domain shift: We assess \textit{FedCC} under a realistic multi-center and multi-device scenario characterized by substantial inter-client variability in scanner type, spatial resolution, image quality, and acquisition conditions. Through extensive comparisons against alternative foundation-model backbones, CNN-based baselines, centralized training, different adaptation strategies, and FL aggregation methods, we show that the proposed LoRA-based federated adaptation improves cross-client generalization and maintains robust performance under heterogeneous acquisition settings.

    \item A novel fetal US dataset for CC detection, acquired during clinical practice across three Italian clinical centers located in Abruzzo, Puglia, and Veneto regions, encompassing images collected with both high- and low-resolution 
    US devices. The dataset captures the substantial variability encountered, and represents a valuable resource for the development and evaluation of generalizable CC detection methods.

\end{itemize}

\section{Related Work}
Only a limited number of studies have specifically addressed the automatic analysis of the CC in fetal US, reflecting both the inherent difficulty of the task and the scarcity of annotated datasets. Among the earliest contributions, \cite{huang2018learning} proposes a framework for the automatic segmentation of the CC and choroid plexus in fetal US images.
The framework first identifies regions with homogeneous intensity patterns and then characterizes them using descriptors that encode both shape and local intensity information. These descriptors are subsequently combined with a boosting classifier to segment the target anatomical structures. 
The framework was evaluated on two small 2-D fetal US datasets: 219 midsagittal images for CC segmentation, acquired from fetuses between 21 and 30 gestational weeks, and 120 trans-thalamic images for choroid plexus segmentation, acquired from healthy fetuses between 20 and 22 gestational weeks. 
The framework achieved high segmentation accuracy, with Dice coefficients of $0.81 \pm 0.06$ for the CC and $0.76 \pm 0.08$ for the choroid plexus (CP). However, its reliance on handcrafted features and a conventional classifier may limit its generalizability across heterogeneous acquisition settings.

More recently, the work in \cite{wang2025fb} proposes FB-ZWUNet, an end-to-end CC segmentation network integrating an attention module for feature enhancement, a wavelet Attention Module for optimized feature fusion, and a morphological constraint module for accurate edge and region capture.
The network was trained on the FB-CC dataset, which consists of 1,336 annotated fetal brain mid-sagittal US images, from 18 to 32 weeks of gestation, collected using three different US devices, achieving a Dice coefficient of 0.874 and IoU of 0.781 on a dedicated fetal brain corpus callosum dataset of 1,336 annotated mid-sagittal US images. The work in \cite{li14deep} proposes CC-FocusNet, a DL framework that integrates automated region localization with an anatomy-aware dual-stream architecture for multi-view analysis of CC abnormalities in prenatal US, achieving 0.973 of accuracy in distinguishing normal from absent CC cases on an independent external test set of 93 cases. 

Although the works in \cite{wang2025fb,li14deep} represent important advances in fetal CC analysis, addressing segmentation and abnormality detection, respectively, both were developed and evaluated within relatively constrained acquisition settings. As a result, their robustness to domain shifts induced by different scanners, imaging protocols, and clinical centers remains unclear. Moreover, neither study considers privacy-preserving collaborative training across institutions, which limits their applicability to realistic multi-center deployment scenarios.

More recently, FL has gained increasing attention in fetal US analysis, demonstrating its potential across a variety of clinical applications while preserving data privacy across institutions. Existing studies have explored FL for tasks ranging from fetal standard plane detection through prototype-based federated denoising approaches \cite{fiorentino2025contrastive}, to fetal biometry and multi-structure abnormality detection with frameworks such as FB-SeUNet++ \cite{judi2026fb}, as well as prenatal congenital heart disease assessment, including the detection of interrupted aortic arch \cite{han2026federated}.

Despite these advances, FL applications in fetal neurosonography remain largely unexplored. In particular, no previous study has investigated CC detection or localization in fetal US within a multi-center, multi-device setting. Furthermore, the potential of FMs adapted through parameter-efficient fine-tuning strategies and trained under privacy-preserving FL paradigms has not yet been examined for this task. This represents a significant gap in the literature, given the clinical importance of CC assessment and the challenges associated with acquiring sufficiently large and diverse neurosonography datasets.

\begin{figure*}[!t]
\centering
    \includegraphics[width=1\linewidth]{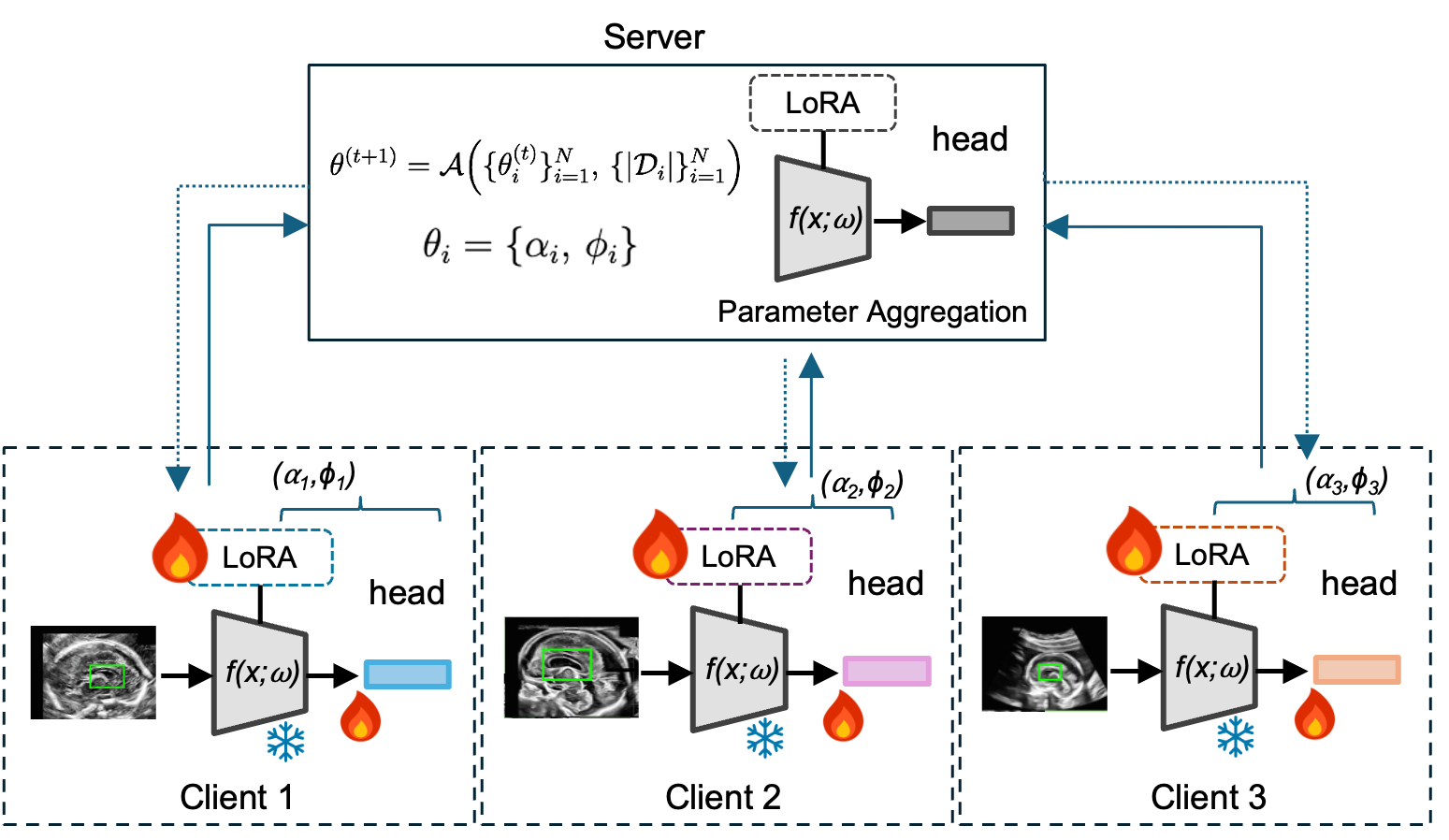}
\caption{Overview of the proposed federated learning framework. Each client trains
a local model consisting of a frozen DINOv2 backbone $f(x;\omega)$, a set of
trainable LoRA adapters $\alpha_i$, and a detection head $\phi_i$. Only the
trainable parameters $\theta_i = \{\alpha_i, \phi_i\}$ are transmitted to the
server, where they are aggregated via a server-side strategy $\mathcal{A}$ and
redistributed at the next communication round. The backbone weights $\omega$ remain
frozen throughout and are never exchanged, substantially reducing communication
overhead while preserving data privacy across clinical sites.}
\label{fig:framework}
\end{figure*}

\section{Materials and methods}
In this section, we first present the overall framework (Sec. \ref{sec:framework}). We then describe the dataset and its federated partitioning (Sec. \ref{sec:dataset}), followed by the configuration adopted for model training (Sec. \ref{training_settings}).

\subsection{Proposed framework}
\label{sec:framework}
The proposed framework, shown in Fig.~\ref{fig:framework}, is built around three integrated components: i) a Vision Transformer (ViT) backbone pretrained through self-supervised learning, ii) a lightweight single-scale detection head derived from the YOLOv8 architecture, and iii) a federated training strategy that coordinates learning across multiple clinical centres without exchanging raw fetal US images.
\\

\textbf{Backbone.} DINOv2 \cite{oquab2023dinov2} is employed as a frozen feature extractor owing to its strong capability to learn transferable representations and its demonstrated robustness in fetal imaging domains. In particular, previous studies have shown that DINO-based representations generalise well across different fetal US datasets and acquisition settings \cite{ambsdorf2025general}. Given an input fetal US image $x \in \mathbb{R}^{H \times W}$, the image is resized and partitioned into non-overlapping patches, which are linearly projected into a sequence of visual tokens and processed by the transformer encoder:
\begin{equation}
z = f_{\mathrm{DINOv2}}(x; \omega),
\label{eq:dinov2}
\end{equation}
where $\omega$ denotes the pretrained DINOv2 parameters and $z$ is the latent representation extracted from the final transformer layer. To preserve the pretrained knowledge encoded in the backbone, the parameters $\omega$ remain frozen throughout training and are never transmitted during federated optimisation.
Adaptation to the CC localisation task is achieved through LoRA modules inserted into the query and value projection matrices of each attention block. Rather than directly updating a pretrained weight matrix $W \in \mathbb{R}^{d \times k}$, LoRA constrains the update to a low-rank residual:
\begin{equation}
W' = W + \Delta W, \qquad \Delta W = BA
\label{eq:lora_update}
\end{equation}
where $A \in \mathbb{R}^{r \times k}$ and $B \in \mathbb{R}^{d \times r}$ are trainable matrices and $r \ll \min(d,k)$ is the LoRA rank. Therefore, the number of trainable parameters is reduced from $d \times k$ to $r(d+k)$.
\\

\begin{figure*}[!t]
\centering
    \includegraphics[width=1\linewidth]{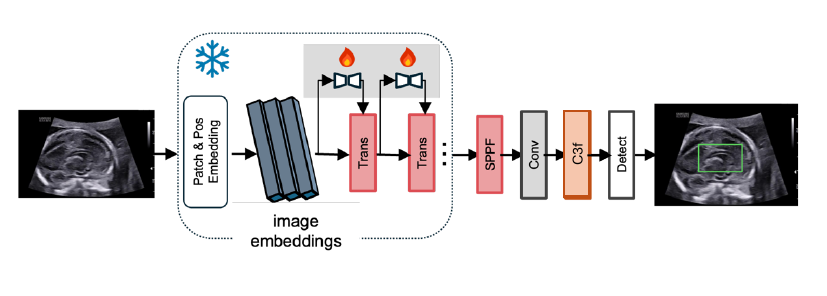}
\caption{Overview of the proposed architecture for corpus callosum localization in ultrasound images. The input image is processed by a frozen DINOv2 backbone, in which LoRA layers are introduced to adapt the Transformer (Trans) features to the target task. The resulting image embeddings are then forwarded to a YOLO-like branch composed of an SPPF and C3f  blocks (\cite{sohan2024review}), followed by the detection head to predict the bounding box.}
\label{fig:detection}
\end{figure*}

\textbf{Detection head.}
To preserve a lightweight and computationally efficient single-stage detection pipeline, the latent representation $z$ extracted by the frozen DINOv2 backbone is reshaped into a two-dimensional feature map and processed by a compact YOLO-like detection branch (Fig. \ref{fig:detection}).
This detection branch enables direct bounding-box regression and confidence prediction, bypassing the need for region proposals or additional refinement stages. Given that the backbone lacks a native multi-resolution feature pyramid, a single detection head was preferred to minimize architectural complexity and facilitate federated optimization. This streamlined approach is further justified by the single-target nature of the task and the localized, compact appearance of the CC. The detection branch can be expressed as

\begin{equation}
h = f_{\mathrm{det}}(z; \phi),
\label{eq:det_head}
\end{equation}
where $\phi$ denotes the trainable parameters of the detection branch and $h$ is the refined feature representation used for localisation. $f_{\mathrm{det}}(\cdot)$ consists of a spatial pyramid pooling--fast (SPPF) block, followed by a convolutional layer, a C3f block, and a final detection layer, following the design principles of YOLOv8 \cite{sohan2024review}. The SPPF, convolutional, and C3f modules were adopted from YOLOv8 because they provide an effective balance between feature refinement and computational efficiency. In particular, the SPPF block enlarges the receptive field and aggregates contextual information at multiple scales, which is beneficial in fetal ultrasound images where the corpus callosum may appear with variable size and surrounding anatomical context. The convolutional layer and C3f block further refine $z$ while maintaining a limited number of trainable parameters and efficient gradient propagation. A detailed description of these modules can be found in \cite{sohan2024review}.

The final detection layer then maps $h$ to the predicted corpus callosum bounding box $\hat{b}$ and corresponding confidence score $\hat{c}$:

\begin{equation}
(\hat{b}, \hat{c}) = g(h; \phi),
\label{eq:bb}
\end{equation}

where $g(\cdot)$ denotes the final prediction layer. This YOLO-inspired design was selected because it provides a favourable trade-off between localisation accuracy and parameter efficiency, which is particularly important in the federated setting, where limiting the number of trainable and shared parameters is essential.
\\

\textbf{Federated training.}
Let $N$ clinical centres participate, indexed by $i \in \{1,\dots,N\}$. The
federally transmitted parameters comprise the LoRA adapters and the detection head
weights:
\begin{equation}    
    \theta_i = \{\alpha_i,\,\phi_i\} = \{A_i,\,B_i,\,\phi_i\},
\label{eq:fl_theta}
\end{equation}

while the backbone $\omega$ remains frozen and is never transmitted. At each
communication round $t$, the server broadcasts the current global parameters
$\theta^{(t)}$ to all selected centres. Each centre minimises a local detection
objective over $E$ epochs:

\begin{equation}
\mathcal{L} = \lambda_{\text{box}} \mathcal{L}_{\text{box}} + \lambda_{\text{cls}} \mathcal{L}_{\text{cls}} + \lambda_{\text{dfl}} \mathcal{L}_{\text{dfl}}
\label{eq:params}
\end{equation}
\\
where $\mathcal{L}{\mathrm{box}}$, $\mathcal{L}{\mathrm{cls}}$, and $\mathcal{L}_{\mathrm{dfl}}$ denote the bounding-box regression, classification, and distribution focal losses, respectively, as defined in the YOLOv8
architecture \cite{Jocher_Ultralytics_YOLO_2023}. Upon completion,
the updated parameters $\theta_i^{(t)}$ are returned to the server and aggregated
via a server-side strategy $\mathcal{A}$:

\begin{equation}    
    \theta^{(t+1)} = \mathcal{A}\!\left(\{\theta_i^{(t)}\}_{i=1}^{N},\,
    \{|\mathcal{D}_i|\}_{i=1}^{N}\right).
\label{eq:server_aggr}
\end{equation}

The updated global parameters are redistributed at the next round. Because neither
raw images, annotations, nor backbone weights are ever exchanged, the approach
substantially reduces communication overhead while preserving data privacy across centres.

\subsection{Dataset}
\label{sec:dataset}
The dataset, collected under the approval of the local Ethical Committees and with the informed consent of all participants, comprises 10,970 fetal US frames extracted from video acquisitions collected from 58 pregnant women during routine neurosonographic examinations. The examinations were acquired at a native frame rate of 25 Hz in three Italian clinical centers: Chieti, Foggia, and Verona. All included cases correspond to normal pregnancies, and no CC malformations were present in the dataset. 
Image acquisition is achieved through the anterior fontanelle of the fetal skull, which provides an acoustic window for US. The CC should be examined in the mid-sagittal plane of the fetal brain, offering a clear view of all its components: splenium, isthmus, genu, corpus, and rostrum. It appears as a thin, elongated, C-shaped structure with two parallel echogenic margins, positioned above the cavum septi pellucidi (CSP)—an anechoic, fluid-filled space that follows the shape of the CC. Below these structures lies the tela choroidea of the third ventricle, seen as a hyperechoic, double-concave line descending toward the cerebellar vermis. The vermis, located at the bottom of the sagittal view, has a hyperechoic shell-like appearance, with an internal tent-shaped indentation representing the fourth ventricle.

Among the collected frames, 4,561 contain a visible CC and were manually annotated by expert clinicians using a single bounding box enclosing the structure. The remaining 6,409 frames do not contain the CC and were retained as background samples to increase the robustness of the detection task. All annotations were provided in YOLO format considering a single target class corresponding to the CC.

To simulate a realistic FL scenario, the dataset was further distributed across three clients, each corresponding to a distinct clinical site and US system. Client 1 included 28 patients and 4,453 frames acquired using a GE Voluson E10 system; Client 2 included 16 patients and 3,391 frames acquired using Samsung HERA W9 and Samsung HERA W10 systems; Client 3 included 14 patients and 3,126 frames acquired using a Samsung HERA Z20 scanner. This partitioning introduces a realistic domain shift across clients, driven by differences in acquisition hardware, imaging settings, and local clinical protocols.

\begin{figure}[!t]
\centering
\includegraphics[width=1\linewidth]{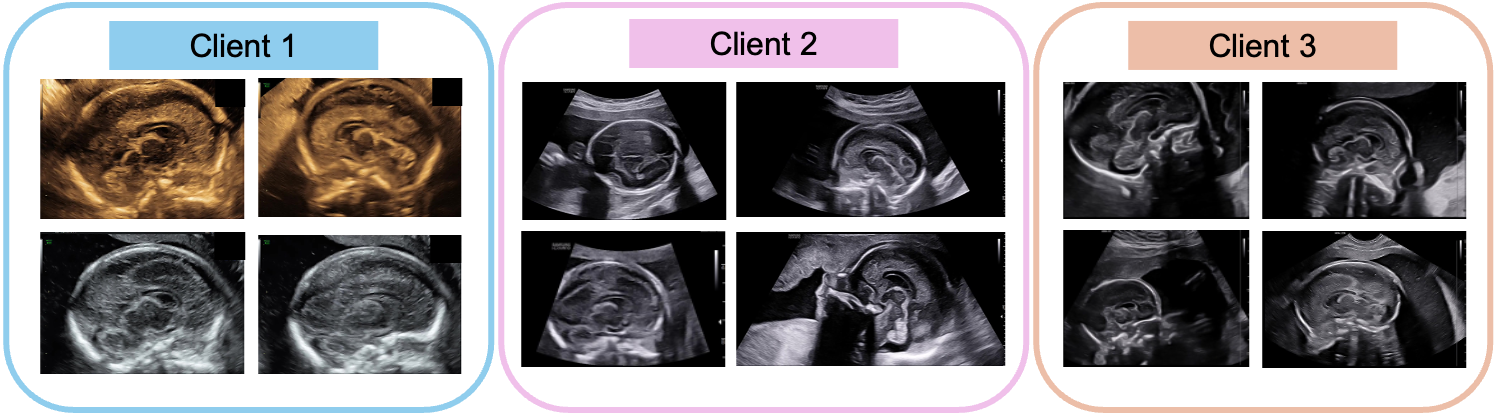}
\caption{Representative fetal ultrasound frames grouped by client, highlighting inter-domain variability across the dataset. The images show differences in field-of-view, fetal pose, and apparent image scale, leading to variability in the size and position of the corpus callosum.}
\label{fig:challenges}
\end{figure}

\begin{table*}[!t]
\centering
\caption{Dataset statistics across federated clients. For each site, 
we report the number of images, the proportion of positive samples 
(CC present), image intensity statistics, spatial resolution, and 
geometric properties of the CC bounding box (normalized area, aspect 
ratio~AR, and vertical center position Ctr-Y). Substantial 
inter-client heterogeneity is observed across all dimensions.}
\label{tab:dataset_stats}
\renewcommand{\arraystretch}{1.3}
\resizebox{\textwidth}{!}{%
\begin{tabular}{lcc cc cc c ccc}
\toprule
& & & \multicolumn{2}{c}{\textbf{Image Statistics}} 
& & \multicolumn{3}{c}{\textbf{BBox Statistics}} \\
\cmidrule(lr){4-5} \cmidrule(lr){7-9}
\textbf{Client} 
& \textbf{\#Patients} 
& \textbf{\#Images} 
& \textbf{Pos.}
& \textbf{Intensity} 
& \textbf{Std} 
& \textbf{Resolution (W$\times$H)}
& \textbf{Area} 
& \textbf{AR} 
& \textbf{Ctr-Y} \\
\midrule
Client 1 & 28 & 4{,}453 & 0.47 
& $71.05 \pm 26.34$ & $67.02 \pm 7.10$ 
& $1278{\pm}219 \times 852$ 
& $0.042{\pm}0.022$ & $1.13{\pm}0.37$ & $0.49{\pm}0.05$ \\

Client 2 & 16 & 3{,}391 & 0.39 
& $58.96 \pm 9.60$ & $70.70 \pm 3.44$ 
& $1062{\pm}348 \times 642{\pm}143$ 
& $0.034{\pm}0.011$ & $1.08{\pm}0.26$ & $0.55{\pm}0.06$ \\

Client 3 & 14 & 3{,}126 & 0.37 
& $60.79 \pm 14.25$ & $68.35 \pm 4.34$ 
& $1232 \times 924$ 
& $0.039{\pm}0.022$ & $1.20{\pm}0.26$ & $0.47{\pm}0.10$ \\
\midrule
\textbf{Total} & \textbf{58} & \textbf{10{,}970} & \textbf{0.41} 
& \multicolumn{6}{c}{---} \\
\bottomrule
\end{tabular}}
\end{table*}

Inter-client heterogeneity is substantial and multi-factorial, as shown in
Fig.~\ref{fig:challenges} and quantified in Table~\ref{tab:dataset_stats}.
Image appearance differs markedly across sites, with well-separated intensity
distributions confirmed by pairwise Wasserstein distances. Spatial resolution is
fixed within Clients 1 and 3, whereas Client 2 exhibits pronounced variability,
reflecting heterogeneous acquisition settings. Geometric properties of the annotated
structures also vary across clients, with systematic differences in the apparent
scale and vertical positioning of the CC. Conversely, bounding-box aspect ratios
remain consistent across sites, suggesting that the morphological appearance of the
CC is largely preserved despite inter-domain variability.

Beyond these statistical differences, the images are affected by challenges intrinsic
to fetal US acquisition, including acoustic shadowing, speckle noise, and
variable image contrast. The apparent position of the CC within the image frame
varies considerably depending on probe orientation and patient-specific factors, as
does the degree of zoom, ranging from close-up views centred on the fetal brain to
wider fields of view encompassing surrounding anatomical structures. In some cases, the CC is only partially visible or poorly delineated, further increasing the
difficulty of the localisation task.

\subsection{Training settings}
\label{training_settings}
Raw US frames were processed through a standardized preprocessing pipeline
prior to training. Scanner-specific overlay artefacts, including textual information
bars and control panels, were removed by zeroing fixed peripheral regions of the
image, preventing the model from exploiting cues unrelated to the target anatomy.

The dataset was divided at the patient level into training, validation, and test subsets for each federated client, preventing data leakage between splits. Client 1 contributed 2,684 training images (15 patients), 644 validation images (3 patients), and 1,125 test images (10 patients). Client 2 contributed 2,031 training images (9 patients), 454 validation images (3 patients), and 906 test images (4 patients), whereas Client 3 contributed 2,165 training images (9 patients), 472 validation images (2 patients), and 683 test images (3 patients). The complete distribution of patients and images across clients and dataset partitions is reported in Table~\ref{tab:data_split}.

\begin{table*}[!t]
\centering
\caption{Patient- and image-level distribution across federated clients and data splits.}
\label{tab:data_split}
\renewcommand{\arraystretch}{1.2}
\footnotesize
\setlength{\tabcolsep}{3pt}
\begin{tabular}{lcccccc}
\toprule
\textbf{Client} 
& \multicolumn{2}{c}{\textbf{Training}} 
& \multicolumn{2}{c}{\textbf{Validation}} 
& \multicolumn{2}{c}{\textbf{Test}} \\
\cmidrule(lr){2-3}\cmidrule(lr){4-5}\cmidrule(lr){6-7}
& \textbf{Patients} & \textbf{Images}
& \textbf{Patients} & \textbf{Images}
& \textbf{Patients} & \textbf{Images} \\
\midrule
Client 1 & 15 & 2{,}684 & 3 & 644 & 10 & 1{,}125 \\
Client 2 & 9  & 2{,}031 & 3 & 454 & 4  & 906 \\
Client 3 & 9 & 1{,}971 & 2 & 445 & 3  & 710 \\
\midrule
\textbf{Total} & \textbf{33} & \textbf{6{,}686} & \textbf{8} & \textbf{1{,}543} & \textbf{17} & \textbf{2{,}741} \\
\bottomrule
\end{tabular}
\end{table*}

Input images were resized to $544 \times 544$ pixels prior to being fed to the
backbone.
All experiments were conducted on NVIDIA A100 GPUs with Automatic Mixed Precision
(AMP) to reduce memory footprint. Each federated simulation runs for 20 communication
rounds with $E = 4$ local epochs per round. Local optimisation was performed using
AdamW with a learning rate of $1\times10^{-4}$, weight decay of $5 \times 10^{-4}$, and a
batch size of 64. The LoRA rank was set to $r = 8$, with scaling factor
$\alpha_{\mathrm{LoRA}} = 16$. The detection loss weights were set to $\lambda_{\mathrm{box}} = 7.5$, $\lambda_{\mathrm{cls}} = 0.5$, and $\lambda_{\mathrm{dfl}} = 1.5$,  with an IoU threshold 
of 0.6 and a confidence threshold of $1\times10^{-3}$. At each communication round,
all $N$ = 3 clients participated in the federated update.


\begin{table*}[t]
\centering
\caption{Summary of the experimental configurations. For each backbone and initialization strategy, all combinations of adaptation strategy and FL aggregation method were evaluated.}
\label{tab:exp_summary}
\resizebox{\textwidth}{!}{%
\begin{tabular}{l l l l l}
\toprule
\textbf{Backbone} & \textbf{Initialization} & \textbf{Detector} & \textbf{Adaptation Strategy} & \textbf{Aggregation} \\
\midrule
$DINOv2_{base}$ & Natural-image pretraining & Single-head / Multi-head & Full FT, Freeze, Proposed & FedAvg, FedProx \\
$DINOv2_{fetal}$ & Fetal US self-supervised pretraining & Single-head & Full FT, Freeze, Proposed & FedAvg, FedProx \\
$DINOv3_{base}$ & Natural-image pretraining & Single-head & Full FT, Freeze, Proposed & FedAvg, FedProx \\
SAM & Natural-image pretraining & Single-head & Full FT, Freeze, Proposed & FedAvg, FedProx \\
UltraSAM & US-specific pretraining & Single-head & Full FT, Freeze, Proposed & FedAvg, FedProx \\
UltraFedFM & Federated US-specific pretraining & Single-head & Full FT, Freeze, Proposed & FedAvg, FedProx \\
\midrule
YOLO26-n & COCO pretraining & Standard detector & Full FT & FedAvg, FedProx \\
YOLO26-s & COCO pretraining & Standard detector & Full FT & FedAvg, FedProx \\
YOLO26-m & COCO pretraining & Standard detector & Full FT & FedAvg, FedProx \\
\bottomrule
\end{tabular}%
}
\end{table*}

\section{Experimental Protocol}
\subsection{Comparisons}
\label{sec:comparisons}

To thoroughly assess the effectiveness of the proposed \textit{FedCC} framework, we designed a comprehensive comparison involving different backbone architectures, pretraining strategies, adaptation schemes, and federated aggregation methods. In particular, the experimental analysis was aimed at answering three main questions: (i) whether, in a federated setting, sharing only parameter-efficient adaptation modules is preferable to updating and sharing the entire backbone; (ii) whether natural-image, generic US, or fetal US-specific pretraining provides the most suitable initialization for fetal CC detection; and (iii) whether a lightweight single-head detector can achieve performance comparable to, or better than, a more complex multi-head architecture.

We considered two variants of the DINOv2 backbone. Both were based on the small architecture (ViT-S/14), which was selected to ensure a fair comparison within the low-resource federated setting addressed in this work, and to prevent performance differences from being driven by model scaling rather than by the proposed adaptation strategy.
The first, denoted as \textbf{$DINOv2_{base}$}, was initialized with the original weights pretrained on large-scale natural image collections \cite{oquab2023dinov2}. The second, referred to as \textbf{$DINOv2_{fetal}$}, used the same architecture but initialized from a version further adapted through self-supervised learning on a publicly available fetal US dataset derived from \cite{conti2025challenging}. Comparing these two variants allows us to isolate the contribution of fetal-specific pretraining with respect to standard natural-image initialization.
Finally, all experiments were performed using both a lightweight single-head detector and its multi-head counterpart. This design allows the effect of the detection architecture to be assessed independently of the backbone initialization and federated optimization strategy, providing a clearer understanding of the contribution of each component within the proposed framework.

To investigate the effect of different FM families, we also considered the Segment Anything Model (SAM), which, although originally pretrained on natural images, has recently been successfully adopted in fetal imaging applications \cite{fiorentino2025adapt, zhou2025segment, le2026viscera}. In addition, we evaluated UltraSAM, an US-specific adaptation of SAM explicitly designed to better model the texture, speckle noise, and anatomical properties of sonographic images \cite{jiang2025ultrasam}. Finally, we considered UltraFedFM \cite{jiang2025pretraining}, a recent privacy-preserving US foundation model pretrained in a federated manner across multiple institutions. UltraFedFM is particularly relevant in this context, as it combines US-specific representations with a training paradigm intrinsically aligned with FL.

For all considered backbones ($DINOv2_{base}$, $DINOv2_{fetal}$, SAM, UltraSAM, and UltraFedFM), we systematically evaluated three adaptation strategies: (i) full fine-tuning (\textit{Full FT}), in which all backbone parameters were updated and shared across clients; (ii) encoder freezing (\textbf{$Freeze$}), in which the pretrained encoder remained fixed and only the detection head was optimized; and (iii) the proposed LoRA-based strategy (\textbf{$proposed$}), in which only a small set of trainable low-rank adapters was introduced and shared among clients. This comparison allows us to quantify whether parameter-efficient adaptation is more effective and communication-efficient than conventional fine-tuning in a federated setting.

Each adaptation strategy was combined with two standard FL aggregation methods, namely FedAvg and FedProx \cite{mcmahan2017communication, li2020federated}. FedAvg was adopted as the reference strategy, as it is the most widely used optimization approach in FL. FedProx was additionally considered to account for the strong heterogeneity expected across fetal US datasets acquired at different institutions, scanners, and gestational ages. By introducing a proximal regularization term during local optimization, FedProx can mitigate client drift and improve convergence under non-IID conditions. Overall, for each backbone and initialization strategy, we evaluated all six possible federated configurations resulting from the combination of the three adaptation schemes and the two aggregation methods.

As representative convolutional baselines, we additionally included three variants of \textbf{$YOLO26$} (nano, small, and medium) \cite{sapkota2025yolo26}, fully fine-tuned from COCO-pretrained weights. These models were selected because the YOLO family is widely recognized for providing a favorable trade-off between detection accuracy and computational efficiency, requiring substantially fewer parameters than large FMs. Including YOLO26 therefore enables us to assess whether the proposed parameter-efficient adaptation strategy can outperform lightweight convolutional models while maintaining a similarly efficient parameter footprint.

\setcounter{equation}{0}

\subsection{Performance settings}
\label{sec:performance_setting}
The detection performance of the proposed framework was evaluated using four standard object detection metrics: Precision, Recall, F1-score, and mean Average Precision at an Intersection over Union (IoU) threshold of 0.5 (mAP@50).

Precision quantifies the fraction of predicted bounding boxes that correctly identify the CC, reflecting the model's tendency to avoid spurious detections. Recall measures the fraction of ground-truth CC instances that are successfully localized, thus characterizing detection sensitivity. The F1-score was computed as their harmonic mean to provide a single balanced summary statistic:
\begin{gather}
    \text{Precision} = \frac{TP}{TP + FP}, \qquad
    \text{Recall}    = \frac{TP}{TP + FN} \\[8pt]
    \text{F1}        = 2 \cdot 
                       \frac{\text{Precision} \cdot \text{Recall}}
                            {\text{Precision} + \text{Recall}}
\end{gather}
where $TP$, $FP$, and $FN$ denote the number of true positives, false positives, and false negatives, respectively.

To obtain a threshold-independent summary of localization 
performance, the Average Precision (AP) was computed as the area under the Precision--Recall curve, and the mAP@50 as its mean across all $N$ evaluated classes:
\begin{equation}
    \text{AP} = \int_{0}^{1} p(r)\,dr, \qquad
    \text{mAP@50} = \frac{1}{N} \sum_{i=1}^{N} 
                    AP_i^{\,\text{IoU}=0.5}
\end{equation}
where $p(r)$ denotes precision as a function of recall. Since the detection task involves a single anatomical target (CC) $N = 1$ and mAP@50 reduces to AP@50. It is nonetheless reported under the standard mAP notation for consistency with the object detection literature \cite{zou2023object}.

In addition, since the proposed framework relies on parameter-efficient adaptation, the different methods were also compared in terms of the number of trainable parameters.

\begin{table*}[!t]
\centering
\caption{
Core comparison study on $DINOv2_{base}$ in the federated setting.
We evaluate adaptation strategy, federated optimization, and initialization.
Metrics are reported per client (C1, C2, C3) and averaged (AVG).
\textbf{Bold} indicates the best result per metric.
}
\label{tab:comparison_core_full}
\resizebox{\textwidth}{!}{%
\begin{tabular}{
ll
c c c >{\columncolor{lightgray}}c
c c c >{\columncolor{lightgray}}c
c c c >{\columncolor{lightgray}}c
c c c >{\columncolor{lightgray}}c
}
\toprule
 & & \multicolumn{4}{c}{\textbf{mAP50}} 
   & \multicolumn{4}{c}{\textbf{Precision}} 
   & \multicolumn{4}{c}{\textbf{Recall}} 
   & \multicolumn{4}{c}{\textbf{F1}} \\
\cmidrule(lr){3-6}\cmidrule(lr){7-10}\cmidrule(lr){11-14}\cmidrule(lr){15-18}
\textbf{Component} & \textbf{Variant} 
  & C1 & C2 & C3 & \cellcolor{lightgray}\textbf{AVG}
  & C1 & C2 & C3 & \cellcolor{lightgray}\textbf{AVG}
  & C1 & C2 & C3 & \cellcolor{lightgray}\textbf{AVG}
  & C1 & C2 & C3 & \cellcolor{lightgray}\textbf{AVG} \\
\midrule
\multicolumn{18}{l}{\textit{Federated -  FedProx}} \\
\midrule
$DINOv2_{fetal}$
& Freeze 
  & 0.670 & 0.961 & 0.472 & 0.701
  & \textbf{0.836} & 0.850 & 0.474 & 0.720
  & 0.584 & 0.954 & 0.559 & 0.699
  & 0.687 & 0.899 & 0.513 & 0.700 \\
  & Full FT 
  & 0.721 & 0.937 & 0.467 & 0.708
  & 0.827 & 0.859 & 0.506 & 0.731
  & 0.660 & 0.939 & 0.504 & 0.701
  & 0.734 & 0.897 & 0.505 & 0.712 \\
  & Proposed 
  & 0.741 & 0.965 & 0.727 & 0.811 
  & 0.827 & 0.949 & 0.654 & 0.810 
  & 0.624 & 0.882 & \textbf{0.757} & 0.754 
  & 0.711 & 0.914 & 0.702 & 0.776 \\
  $DINOv2_{base}$
& Freeze 
  & 0.705 & 0.961 & 0.548 & 0.738
  & 0.713 & 0.912 & 0.554 & 0.726
  & 0.684 & 0.903 & 0.593 & 0.727
  & 0.698 & 0.907 & 0.573 & 0.726 \\
  & Full FT 
  & 0.742 & 0.969 & 0.601 & 0.771
  & 0.735 & 0.935 & 0.622 & 0.764
  & 0.702 & 0.926 & 0.641 & 0.756
  & 0.718 & \textbf{0.930} & 0.631 & 0.760 \\
  & Proposed
  & \textbf{0.814} & 0.973 & 0.747 & 0.845
  & 0.752 & 0.922 & 0.683 & 0.786
  & 0.739 & 0.901 & 0.683 & 0.774
  & 0.752 & 0.922 & 0.683 & 0.786 \\
\midrule
\multicolumn{18}{l}{\textit{Federated - FedAvg}} \\
\midrule
$DINOv2_{fetal}$
& Freeze 
  & 0.711 & 0.931 & 0.433 & 0.692
  & 0.812 & 0.854 & 0.457 & 0.708
  & 0.660 & 0.935 & 0.485 & 0.694
  & 0.728 & 0.893 & 0.471 & 0.697 \\
  & Full FT 
  & 0.688 & 0.970 & 0.507 & 0.722
  & 0.830 & 0.857 & 0.632 & 0.773
  & 0.617 & \textbf{0.973} & 0.441 & 0.677
  & 0.708 & 0.912 & 0.520 & 0.713 \\
  & Proposed 
  & 0.741 & \textbf{0.978} & \textbf{0.800} & 0.839 
  & 0.705 & 0.892 & 0.780 & 0.792 
  & 0.672 & 0.966 & 0.731 & \textbf{0.790} 
  & 0.688 & 0.927 & \textbf{0.755} & 0.790 \\
  $DINOv2_{base}$
& Freeze 
  & 0.681 & 0.957 & 0.507 & 0.715
  & 0.692 & 0.905 & 0.516 & 0.704
  & 0.657 & 0.892 & 0.564 & 0.704
  & 0.674 & 0.898 & 0.539 & 0.704 \\
  & Full FT 
  & 0.630 & 0.953 & 0.579 & 0.721
  & 0.706 & 0.924 & 0.569 & 0.733
  & 0.671 & 0.914 & 0.615 & 0.733
  & 0.688 & 0.924 & 0.587 & 0.733 \\
  & Proposed
  & 0.806 & 0.977 & 0.788 & \textbf{0.857}
  & 0.761 & \textbf{0.957} & \textbf{0.846} & \textbf{0.855}
  & \textbf{0.746} & 0.890 & 0.646 & 0.761
  & \textbf{0.753} & 0.922 & 0.732 & \textbf{0.803} \\
\bottomrule
\end{tabular}%
}
\end{table*}

\begin{table*}[t]
\centering
\caption{
Impact of detection head design for $DINOv2_{base}$ pretrained.
We compare single-scale and multi-scale (2-scale, 3-scale) detection heads using
their best-performing configurations. Metrics are reported per client (C1, C2, C3)
and averaged (AVG). \textbf{Bold} indicates the best result per metric.
}
\label{tab:comparison_dete}
\resizebox{\textwidth}{!}{%
\begin{tabular}{
ll
c c c >{\columncolor{lightgray}}c
c c c >{\columncolor{lightgray}}c
c c c >{\columncolor{lightgray}}c
c c c >{\columncolor{lightgray}}c
}
\toprule
 & & \multicolumn{4}{c}{\textbf{mAP50}} 
   & \multicolumn{4}{c}{\textbf{Precision}} 
   & \multicolumn{4}{c}{\textbf{Recall}} 
   & \multicolumn{4}{c}{\textbf{F1}} \\
\cmidrule(lr){3-6}\cmidrule(lr){7-10}\cmidrule(lr){11-14}\cmidrule(lr){15-18}
\textbf{Component} & \textbf{Variant} 
  & C1 & C2 & C3 & AVG 
  & C1 & C2 & C3 & AVG 
  & C1 & C2 & C3 & AVG 
  & C1 & C2 & C3 & AVG \\
\midrule

\multicolumn{18}{l}{\textit{Detection Head}} \\
\midrule
  $DINOv2_{base}$
  & 3-scale
  & 0.805 & 0.973 & 0.445 & 0.741
  & 0.729 & 0.913 & 0.482 & 0.708
  & 0.729 & 0.913 & 0.482 & 0.708
  & 0.729 & 0.913 & 0.482 & 0.708 \\

  & 2-scale
  & 0.786 & 0.965 & 0.520 & 0.757
  & 0.721 & 0.918 & 0.517 & 0.719
  & 0.721 & 0.918 & 0.517 & 0.719
  & 0.721 & 0.918 & 0.517 & 0.719 \\
  & \textbf{Single-scale}
  & \textbf{0.806} & \textbf{0.977} & \textbf{0.788} & \textbf{0.857}
  & \textbf{0.761} & \textbf{0.957} & \textbf{0.846} & \textbf{0.855}
  & \textbf{0.746} & 0.890 & \textbf{0.646} & \textbf{0.761}
  & \textbf{0.753} & \textbf{0.922} & \textbf{0.732} & \textbf{0.803} \\

\bottomrule
\end{tabular}%
}
\end{table*}
\begin{table*}[!ht]
\centering
\caption{Federated performance of the considered FM-based backbones under FedAvg and FedProx. Best result for each metric is highlighted in bold.}
\label{tab:fed_cen_results}
\resizebox{\textwidth}{!}{%
\begin{tabular}{
ll
c c c >{\columncolor{lightgray}}c
c c c >{\columncolor{lightgray}}c
c c c >{\columncolor{lightgray}}c
c c c >{\columncolor{lightgray}}c
}
\toprule
& & \multicolumn{4}{c}{mAP50} & \multicolumn{4}{c}{Precision} & \multicolumn{4}{c}{Recall} & \multicolumn{4}{c}{F1} \\
\cmidrule(lr){3-6} \cmidrule(lr){7-10} \cmidrule(lr){11-14} \cmidrule(lr){15-18}
\textbf{Model} & \textbf{Tuning}
& C1 & C2 & C3 & \cellcolor{lightgray}\textbf{AVG}
& C1 & C2 & C3 & \cellcolor{lightgray}\textbf{AVG}
& C1 & C2 & C3 & \cellcolor{lightgray}\textbf{AVG}
& C1 & C2 & C3 & \cellcolor{lightgray}\textbf{AVG} \\
\midrule
\multicolumn{18}{l}{\textit{Centralized}} \\
\midrule
YOLO26n & Full FT 
& 0.777 & 0.964 & 0.165 & 0.635 
& 0.752 & 0.937 & 0.199 & 0.630 
& 0.679 & 0.863 & 0.301 & 0.614 
& 0.7140 & 0.8986 & 0.2402 & 0.617 \\

YOLO26s & Full FT 
& 0.749 & 0.941 & 0.561 & 0.750 
& 0.766 & 0.951 & 0.601 & 0.773 
& 0.660 & 0.882 & 0.555 & 0.699 
& 0.709 & 0.915 & 0.577 & 0.734 \\

YOLO26m & Full FT 
& 0.622 & 0.921 & 0.549 & 0.697 
& 0.710 & 0.903 & 0.586 & 0.733 
& 0.550 & 0.856 & 0.521 & 0.642 
& 0.620 & 0.879 & 0.552 & 0.684 \\

$DINOv3_{base}$ & Full FT 
& 0.732 & 0.955 & 0.636 & 0.774 
& 0.776 & 0.884 & 0.704 & 0.788 
& 0.677 & 0.939 & 0.552 & 0.723 
& 0.723 & 0.911 & 0.619 & 0.751 \\

$DINOv2_{base}$ & proposed 
& 0.724 & 0.977 & 0.422 & 0.708 
& 0.785 & 0.952 & 0.512 & 0.750 
& 0.661 & 0.943 & 0.478 & 0.694 
& 0.718 & \textbf{0.947} & 0.495 & 0.720 \\

$DINOv2_{fetal}$ & proposed 
& 0.752 & \textbf{0.983} & 0.608 & 0.781 
& 0.797 & \textbf{0.956} & 0.633 & 0.795 
& 0.645 & 0.920 & 0.537 & 0.701 
& 0.713 & 0.938 & 0.581 & 0.744 \\
\midrule
\multicolumn{18}{l}{\textit{Federated -- FedProx}} \\
\midrule

YOLO26n & Full FT & 0.788 & 0.959 & 0.673 & 0.807 & 0.854 & 0.908 & 0.759 & 0.840 & 0.698 & 0.939 & 0.601 & 0.746 & 0.769 & 0.923 & 0.670 & 0.787 \\

YOLO26s         & Full FT   & 0.834 & 0.975 & 0.518 & 0.775  & 0.859 & 0.941 & 0.563 & 0.787  & 0.801 & 0.901 & 0.493 & 0.732  & 0.829 & 0.921 & 0.525 & 0.758 \\
YOLO26m         & Full FT   & 0.791 & 0.966 & 0.734 & 0.830  & 0.832 & 0.923 & 0.763 & 0.839  & 0.728 & 0.875 & 0.728 & 0.777  & 0.777 & 0.898 & 0.728 & 0.801 \\
SAM & proposed & 0.689 & 0.952 & 0.448 & 0.696 & 0.784 & 0.903 & 0.472 & 0.720 & 0.590 & 0.889 & 0.552 & 0.677 & 0.673 & 0.896 & 0.508 & 0.692 \\
$DINOv3_{base}$ & Feat.Ext & 0.788 & 0.974 & 0.711 & 0.824 & 0.818 & 0.908 & 0.662 & 0.796 & 0.666 & 0.962 & 0.676 & 0.768 & 0.734 & 0.934 & 0.669 & 0.779 \\
UltraFedFM & Full FT & 0.675 & 0.969 & 0.243 & 0.629 & 0.858 & 0.845 & 0.359 & 0.687 & 0.590 & 0.977 & 0.449 & 0.672 & 0.699 & 0.906 & 0.399 & 0.668 \\
UltraSAM & Full FT & 0.796 & 0.979 & 0.614 & 0.796 & 0.808 & \textbf{0.973} & 0.542 & 0.775 & 0.683 & 0.901 & 0.610 & 0.732 & 0.741 & 0.936 & 0.574 & 0.750 \\
$DINOv2_{fetal}$ & proposed & 0.741 & 0.965 & 0.727 & 0.811 & 0.827 & 0.949 & 0.654 & 0.810 & 0.624 & 0.882 & \textbf{0.757} & 0.754 & 0.711 & 0.914 & 0.702 & 0.776 \\
$DINOv2_{base}$& Proposed
  & 0.814 & 0.973 & 0.747 & 0.845
  & 0.752 & 0.922 & 0.683 & 0.786
  & 0.739 & 0.901 & 0.683 & 0.774
  & 0.752 & 0.922 & 0.683 & 0.786 \\
\midrule
\multicolumn{18}{l}{\textit{Federated -- FedAvg}} \\
\midrule
YOLO26n         & Full FT   & 0.790 & 0.972 & 0.594 & 0.785  & 0.877 & 0.930 & 0.637 & 0.815  & 0.678 & 0.970 & 0.566 & 0.738  & 0.765 & 0.949 & 0.599 & 0.771 \\
YOLO26s         & Full FT   & 0.841 & 0.960 & 0.458 & 0.753  & 0.892 & 0.913 & 0.439 & 0.748  & 0.731 & 0.932 & 0.599 & 0.754  & 0.804 & 0.922 & 0.507 & 0.744 \\
YOLO26m         & Full FT   & 0.802 & 0.977 & 0.416 & 0.732  & 0.869 & 0.967 & 0.457 & 0.764  & 0.727 & 0.890 & 0.515 & 0.710  & 0.791 & 0.9266 & 0.4845 & 0.7342 \\
SAM & proposed & 0.673 & 0.955 & 0.427 & 0.685 & 0.778 & 0.899 & 0.454 & 0.710 & 0.567 & 0.894 & 0.538 & 0.666 & 0.656 & 0.896 & 0.492 & 0.681 \\
$DINOv3_{base}$ & proposed & 0.773 & 0.972 & 0.724 & 0.823 & 0.799 & 0.945 & 0.707 & 0.817 & 0.656 & 0.909 & 0.690 & 0.752 & 0.721 & 0.927 & 0.698 & 0.782 
\\UltraFedFM & proposed & 0.652 & 0.965 & 0.176 & 0.598 & 0.768 & 0.904 & 0.279 & 0.650 & 0.550 & 0.920 & 0.471 & 0.647 & 0.641 & 0.912 & 0.350 & 0.634 \\
UltraSAM & Full FT & \textbf{0.834} & 0.980 & 0.742 & 0.852 & \textbf{0.877} & 0.960 & 0.636 & 0.824 & 0.678 & 0.901 & 0.720 & 0.767 & \textbf{0.765} & 0.930 & 0.676 & 0.790 \\
$DINOv2_{fetal}$ & proposed & 0.741 & 0.978 & \textbf{0.800} & 0.839 & 0.705 & 0.892 & 0.780 & 0.792 & 0.672 & \textbf{0.966} & 0.731 & \textbf{0.790} & 0.688 & 0.927 & \textbf{0.755} & 0.790 \\
$DINOv2_{base}$ & proposed & 0.806 & 0.977 & 0.788 & \textbf{0.857} & 0.761 & 0.957 & \textbf{0.846} & \textbf{0.855} & \textbf{0.746} & 0.890 & 0.646 & 0.761 & 0.753 & 0.922 & 0.732 & \textbf{0.803} \\

\bottomrule
\end{tabular}%
}
\end{table*}
\begin{figure}[!t]
\centering
\hspace*{-0.5cm}
\includegraphics[width=1.05\linewidth]{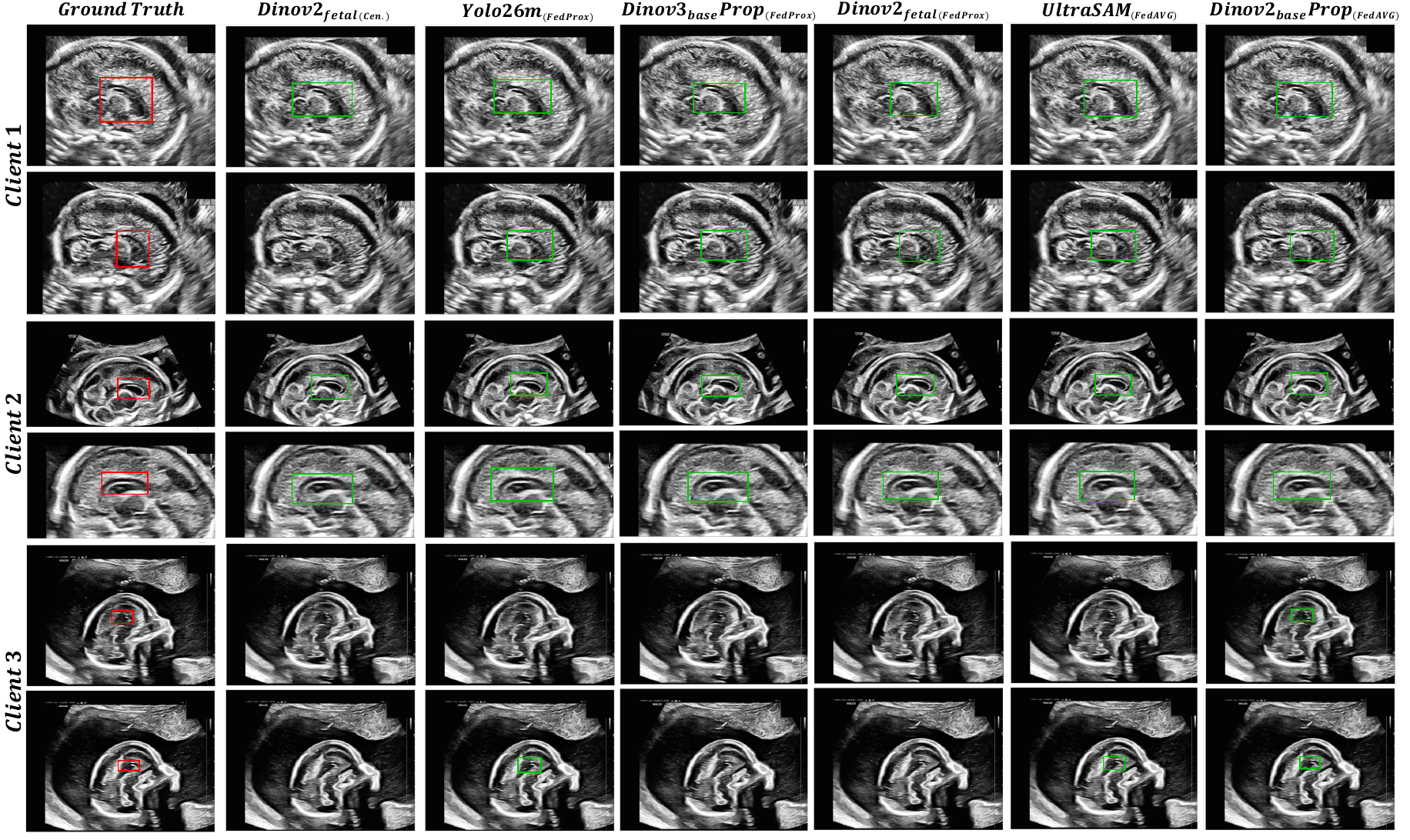}
\caption{Qualitative comparison of corpus callosum localization across three federated clients and six models: $DINOv2_{fetal}$ (Centralized), YOLO26m (FedProx), $DINOv3_{base}$ Proposed (FedProx), $DINOv2_{fetal}$ (FedAvg), UltraSAM (FedAvg), and $DINOv2_{base}$ Proposed (FedAvg). Ground truth annotations are shown in red; model predictions in green. Each row pair corresponds to one federated client, reflecting distinct acquisition domains. 
}
\label{fig:qualitative}
\end{figure}
\begin{figure}[!t]
\centering
\includegraphics[width=1\linewidth]{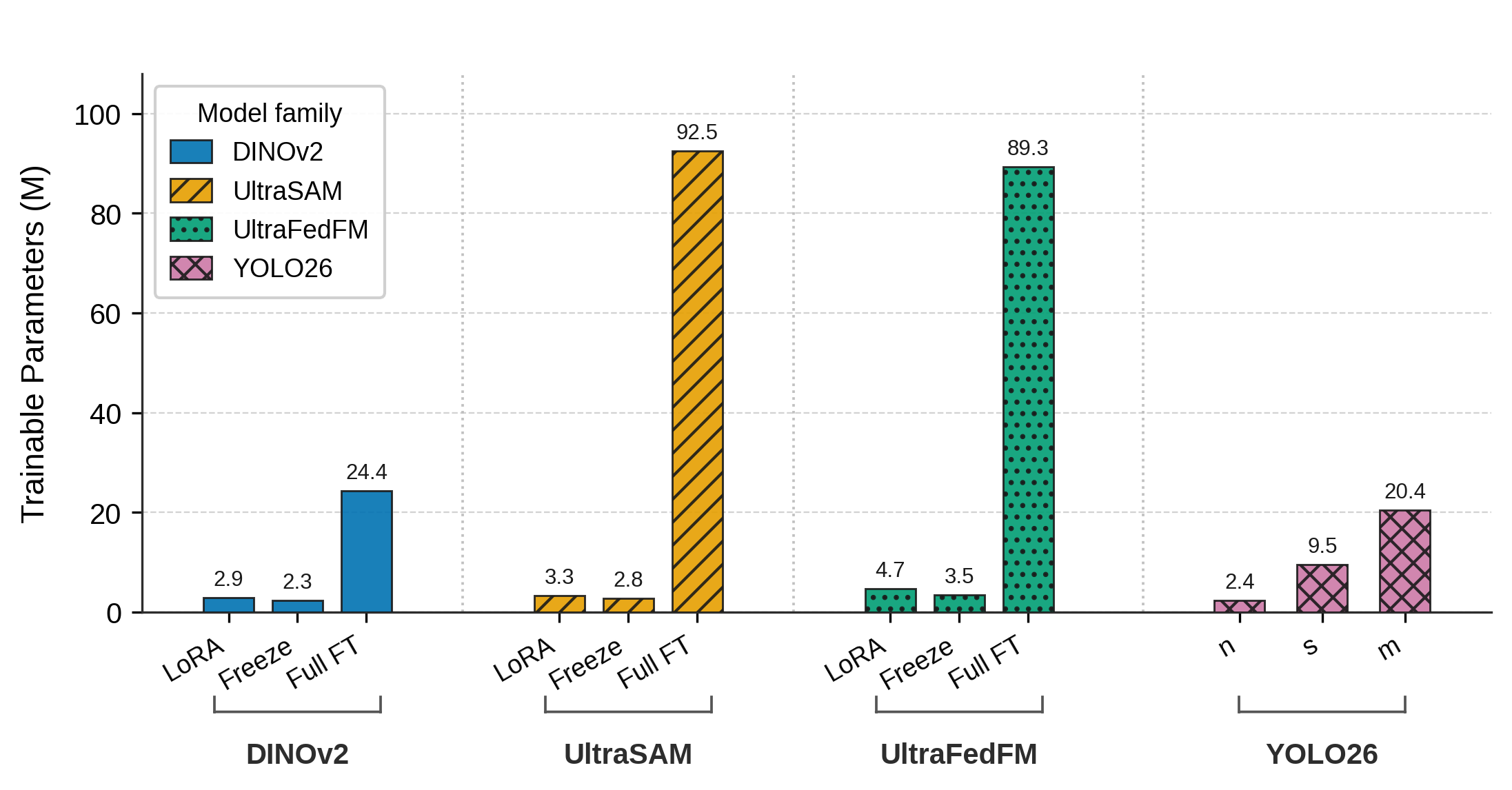}
\caption{Trainable parameter count across model families and adaptation strategies. Each bar reports the number of trainable parameters (in millions, M) for a given combination of backbone architecture and fine-tuning strategy. Four model families are compared: DINOv2, UltraSAM, UltraFedFM, and YOLO26 (variants n/s/m). For the first three families, three adaptation strategies are evaluated: Low-Rank Adaptation (LoRA), encoder freezing with trainable decoder (Freeze), and full fine-tuning (Full FT). Parameter counts are reported as floating-point millions. Hatch patterns distinguish model families to ensure readability in greyscale reproduction.}
\label{fig:params}
\end{figure}

\section{Results}
\label{sec:results}
Table \ref{tab:comparison_core_full} reports the comparison between the two DINOv2 initializations, $DINOv2_{base}$ and $DINOv2_{fetal}$, under different adaptation strategies and federated optimization methods.
Across both FedAvg and FedProx, $DINOv2_{base}$ consistently achieves higher average performance compared to $DINOv2_{fetal}$. Under FedAvg, the best results are obtained with the proposed adaptation strategy applied to $DINOv2_{base}$, reaching an average mAP50 of 0.857 and an F1-score of 0.803. Similarly, under FedProx, the same configuration achieves the highest performance, with an average mAP50 of 0.845 and F1-score of 0.786.
For $DINOv2_{fetal}$, the proposed strategy improves performance compared to both freezing and full FT across most metrics and clients. Under FedAvg, it achieves an average mAP50 of 0.839 and F1-score of 0.790, while under FedProx it reaches 0.811 and 0.776, respectively.
Given the superior performance of $DINOv2_{base}$ across all settings, Table~\ref{tab:comparison_dete} further investigates this configuration by analyzing the impact of the detection head design. The single-scale configuration achieves the highest performance across all metrics, with an average mAP50 of 0.857 and F1-score of 0.803. The 2-scale and 3-scale variants show lower results, with average mAP50 values of 0.757 and 0.741, respectively.

Table~\ref{tab:fed_cen_results} reports the performance of all considered FM-based backbones, comparing centralized training with federated optimization under both FedProx and FedAvg.
In the centralized setting, YOLO-based models achieve competitive performance, with YOLO26s reaching the highest average mAP50 (0.750) and F1-score (0.734). Among transformer-based models, $DINOv2_{fetal}$ slightly outperforms $DINOv2_{base}$, achieving an average mAP50 of 0.781 and F1-score of 0.744 when combined with the proposed adaptation strategy.
Under federated optimization, all models show an overall improvement compared to the centralized setting. In particular, under FedAvg, $DINOv2_{base}$ with the proposed strategy achieves the best overall performance, reaching an average mAP50 of 0.857 and F1-score of 0.803. Similarly, $DINOv2_{fetal}$ achieves strong results, with an average mAP50 of 0.839 and F1-score of 0.790.
Among the alternative FMs, UltraSAM and UltraFedFM show competitive performance, especially under FedAvg. UltraSAM achieves an average F1-score of 0.790, comparable to $DINOv2_{fetal}$, while UltraFedFM shows lower performance overall, particularly in terms of mAP50.
SAM exhibits lower performance compared to both DINOv2 variants and UltraSAM in all configurations. Similarly, YOLO-based models, despite strong centralized results, do not reach the same performance levels in the federated setting.
Under FedProx, a similar trend is observed. $DINOv2_{base}$ with the proposed strategy achieves the highest average performance (mAP50 = 0.845, F1 = 0.786), followed by $DINOv2_{fetal}$ (mAP50 = 0.811, F1 = 0.776). Other models, including UltraSAM and YOLO variants, achieve lower but comparable results.

Figure~\ref{fig:qualitative} presents qualitative results across the three federated clients for a representative set of models: $DINOv2_{fetal}$ (Centralized), YOLO26m (FedProx), $DINOv3_{base}$ with the proposed strategy (FedProx), $DINOv2_{fetal}$ (FedAvg) with the proposed strategy, UltraSAM (FedAvg), and $DINOv2_{base}$ with the proposed strategy (FedAvg). Ground truth annotations are shown in red, while model predictions are shown in green. Client 1 presents several inherently challenging examples in which the CC is only partially visible or exhibits low contrast against the surrounding fetal brain tissue. 
However, most of the models exhibits good results.
The proposed LoRA-based strategies demonstrate comparatively more stable predictions, better aligned with the ground truth even under low-visibility conditions. Client 2 represents the least challenging domain in the federation. The CC is generally well-delineated and clearly distinguishable from adjacent anatomical structures, resulting in consistent and accurate predictions across most models. This client-level advantage is reflected in the quantitative results, where all methods achieve their highest per-client scores. Client 3 introduces a distinct type of difficulty: while the CC is frequently present within the field of view, its boundaries tend to merge with surrounding fetal soft tissue, reducing edge contrast and increasing anatomical ambiguity. This domain shift, driven by differences in scanner characteristics, fetal positioning, or gestational stage, visibly degrades the predictions of several baseline models, which either miss the structure entirely or produce poorly localized outputs. The proposed federated adaptation methods, by contrast, retain greater localization accuracy, suggesting improved robustness to appearance variability across acquisition domains.

Figure~\ref{fig:params} reports the number of trainable parameters across the different model families and adaptation strategies. For all FMs, the proposed LoRA-based approach drastically reduces the number of trainable parameters compared with full FT, while remaining close to the freezing configuration. Specifically, for DINOv2, LoRA requires only 2.9M trainable parameters, compared with 24.4M for full FT and 2.3M when the encoder is frozen. A similar behavior is observed for UltraSAM, where LoRA involves 3.3M trainable parameters versus 92.5M for full FT, and for UltraFedFM, with 4.7M parameters compared with 89.3M under the full FT setting. Overall, the proposed strategy consistently achieves a substantial reduction in trainable parameters across all considered backbones, offering a parameter-efficient alternative. In contrast, the number of trainable parameters in the YOLO26 models varies according to model scale, ranging from 2.4M for the nano variant to 20.4M for the medium version.

\section{Discussion}
\label{sec:discussion}

Automatic detection of CC in fetal neurosonography is a clinically relevant yet challenging task, requiring models that are robust to scanner-dependent variability, effective under limited supervision, and compatible with privacy-preserving clinical deployment. 
The results reported in Tables~\ref{tab:comparison_core_full}, \ref{tab:fed_cen_results}, and \ref{tab:comparison_dete} indicate that these requirements can be effectively addressed by combining a self-supervised transformer backbone with parameter-efficient adaptation in a federated setting.
The comparison between $DINOv2_{base}$ and $DINOv2_{fetal}$ provides an interesting insight into the role of domain-specific pretraining in federated fetal US analysis. Although fetal-specific initialization could be expected to provide a better starting point for CC detection, $DINOv2_{base}$ consistently achieved higher average performance across both FedAvg and FedProx. This result suggests that, in this setting, the breadth and diversity of the original DINOv2 pretraining may be more beneficial than a subsequent domain-specific adaptation performed on a substantially smaller fetal US dataset. A possible explanation is that large-scale self-supervised pretraining on highly diverse natural images allows $DINOv2_{base}$ to learn general visual representations that remain transferable to fetal US, especially when coupled with a parameter-efficient adaptation strategy. Conversely, the fetal-pretrained variant may have shifted the representation toward the characteristics of the pretraining fetal dataset, potentially reducing its generality when applied to CC data. This effect may be particularly relevant in FL, where each client can exhibit scanner-, protocol-, and population-dependent appearance differences. Therefore, while domain-specific pretraining remains a promising direction, these results indicate that its effectiveness depends not only on domain similarity, but also on the scale, diversity, and representativeness of the pretraining data. 

The results also show that the proposed LoRA-based adaptation consistently provides the best or near-best performance for both initializations and aggregation strategies. This suggests that updating a limited set of trainable parameters can better preserve the transferable representations learned during pretraining while still enabling task-specific adaptation. In contrast, full FT does not consistently improve performance and, in some cases, leads to lower results, supporting the use of parameter-efficient adaptation in low-resource and heterogeneous federated settings.
The comparison with other FM backbones further highlights differences in performance across architectures. In particular, SAM consistently achieves lower results across all configurations. A possible explanation is that SAM is primarily designed as a prompt-based segmentation model, where performance depends on the availability and quality of input prompts. In the considered detection setting, where no explicit prompt information is provided, its representations may be less aligned with the task requirements. In addition, its pretraining on natural images may limit its ability to capture the specific appearance characteristics of fetal US data. UltraSAM represents the closest FM-based competitor, achieving performance comparable to the proposed approach under FedAvg. However, its performance decreases more markedly on the most challenging client (C3), where both mAP50 and F1-score drop compared to $DINOv2_{base}$. This behavior may be related to the higher model capacity and the increased number of trainable parameters, which could make the model more sensitive to limited data availability and domain shifts across clients. Similarly, UltraFedFM does not achieve performance comparable to DINOv2-based models. Although it benefits from US-specific pretraining, its relatively large number of trainable parameters may limit its robustness in the considered federated setting, where each client provides only a subset of the overall data distribution. This can result in less stable generalization across heterogeneous centers. 
Compared with CNN-based YOLO26 baselines, $DINOv2$ exhibits more consistent performance across clients. This difference is unlikely to be explained by model capacity alone, as YOLO26m has a comparable number of parameters. Instead, it suggests that the self-supervised representations learned by DINOv2 are more transferable under scanner-dependent appearance shifts. This is consistent with the design of DINOv2, which leverages large-scale self-supervised pretraining to learn general-purpose visual features~\cite{oquab2023dinov2}, and with prior evidence highlighting the impact of scanner, vendor, and protocol variability on model generalization in US imaging~\cite{fiorentino2023review,torres2022review}.

The comparison between FedAvg and FedProx should be interpreted in a model-dependent manner. FedProx was introduced to improve federated optimization under statistical and systems heterogeneity by constraining local updates toward the global model~\cite{li2020federated}. In our experiments, this proximal regularization improved some configurations, most notably the CNN-based YOLO26m baseline, whose average mAP50 increased from 0.732 under FedAvg to 0.830 under FedProx. This suggests that models trained through full fine-tuning may benefit from an explicit constraint on local optimization when client distributions are heterogeneous. However, the same trend was not observed for the proposed LoRA-based DINOv2 configuration. One possible explanation is that LoRA already constrains local adaptation by limiting trainable updates to a compact low-rank parameter space. In this setting, the additional proximal term may provide limited extra regularization and may partially reduce the flexibility required to adapt to the smallest and most shifted client. This interpretation is consistent with prior observations that client heterogeneity, data imbalance, and client drift can strongly influence federated optimization dynamics~\cite{mcmahan2017communication, karimireddy2020scaffold}. From a deployment perspective, the fact that FedAvg provided the best trade-off for the proposed LoRA-based configuration is practically relevant. FedAvg avoids the need to tune the proximal coefficient required by FedProx, reduces local optimization complexity, and simplifies deployment across heterogeneous clinical infrastructures. This advantage is further reinforced by the communication efficiency of the proposed approach. 
The proposed federated configuration outperformed its centralized counterpart trained on pooled data. Although centralized training is often considered an upper-bound setting because all data are available during optimization, this assumption does not necessarily hold when the pooled dataset is heterogeneous and imbalanced across domains. In the present study, centralized $DINOv2_{base}$ achieved an average mAP50 of 0.708 and degraded substantially on C3, reaching only 0.422 mAP50. By contrast, the federated $DINOv2_{base}$ + LoRA model trained with FedAvg achieved 0.857 average mAP50 and retained 0.788 mAP50 on C3. This suggests that federated optimization may have acted as a form of structured multi-domain training, where each client contributes domain-specific updates before aggregation. In this setting, the repeated local training and aggregation cycle may help the model preserve representations that are useful across scanners, rather than overfitting to the dominant acquisition domain in the pooled dataset. However, this interpretation should be considered specific to the dataset and partitioning used in this study, and should be further validated on larger multi-center cohorts.

Several limitations must be acknowledged. First, although the dataset is multi-center and multi-device, it includes 58 patients from three clinical sites. Validation on larger and more diverse cohorts, including additional centers, US manufacturers, acquisition protocols, and gestational-age ranges, is necessary before considering clinical translation \cite{fiorentino2025uncovering}. Second, all included cases correspond to normal pregnancies, and no CC malformations were present. Therefore, the current study evaluates localization robustness but does not address abnormality detection or classification. Third, the framework performs bounding-box detection rather than segmentation, preventing direct extraction of biometric measurements such as CC length, thickness, area, or subregional morphology. Fourth, the federated setting was simulated under controlled conditions, with all clients participating at each communication round. Real cross-institutional deployments may involve asynchronous updates, client dropout.

\section{Conclusions}
\label{sec:conclusions}

In this work, we proposed a federated and parameter efficient framework for automated detection of CC in fetal neurosonography, addressing the joint challenges of privacy preservation, inter-scanner domain shift, and communication efficiency. Our central contribution is the combination of a DINOv2 backbone with LoRA in a federated pipeline, which we evaluated on a multi-site dataset of 10,970 ultrasound frames from 58 patients acquired with four distinct scanners.

Our results suggest that LoRA serves a dual role in this context: it reduces the per-round parameter transmission by $8.5\times$ and provides a constrained adaptation space that may improve the stability of federated training under heterogeneous client distributions.

Future work will extend this framework toward automated CC segmentation for biometric quantification and validation on larger and more diverse fetal US datasets collected across multiple clinical centers. Expanding the number of participating institutions and increasing dataset heterogeneity will enable a more comprehensive assessment of model robustness and generalizability across different acquisition settings and patient populations. In addition, a key future direction will be the detection of other anatomical subregions of the CC, such as the rostrum and splenium, as accurate localization of these structures may provide valuable structural and developmental information~\cite{cha2022altered,lubian2024corpus}.

\section{Aknowledgements}

This work was partially supported  by the Italian Fund
for Applied Sciences (FISA), grant no. FISA2022-00696.

\section{Code Availability}
The source code will be made publicly available upon acceptance of the manuscript.

\section{Supplementary Material}
\label{app:supplementary}

\begin{landscape}
\scriptsize
\setlength{\tabcolsep}{2pt}
\renewcommand{\arraystretch}{1.08}
\begin{longtable}{
ll
c c c c
c c c c
c c c c
c c c c
}
\caption{Expanded comparison of the experimental configurations associated with the main comparison study. Metrics are reported per client (C1, C2, C3) and averaged (AVG). Only configurations directly related to the models and settings discussed in the main comparison are included.}
\label{tab:supplementary_expanded_comparison}\\
\toprule
& & \multicolumn{4}{c}{\textbf{mAP50}}
& \multicolumn{4}{c}{\textbf{Precision}}
& \multicolumn{4}{c}{\textbf{Recall}}
& \multicolumn{4}{c}{\textbf{F1}} \\
\cmidrule(lr){3-6}\cmidrule(lr){7-10}\cmidrule(lr){11-14}\cmidrule(lr){15-18}
\textbf{Model} & \textbf{Tuning}
& C1 & C2 & C3 & AVG
& C1 & C2 & C3 & AVG
& C1 & C2 & C3 & AVG
& C1 & C2 & C3 & AVG \\
\midrule
\endfirsthead
\toprule
& & \multicolumn{4}{c}{\textbf{mAP50}}
& \multicolumn{4}{c}{\textbf{Precision}}
& \multicolumn{4}{c}{\textbf{Recall}}
& \multicolumn{4}{c}{\textbf{F1}} \\
\cmidrule(lr){3-6}\cmidrule(lr){7-10}\cmidrule(lr){11-14}\cmidrule(lr){15-18}
\textbf{Model} & \textbf{Tuning}
& C1 & C2 & C3 & AVG
& C1 & C2 & C3 & AVG
& C1 & C2 & C3 & AVG
& C1 & C2 & C3 & AVG \\
\midrule
\endhead
\midrule
\multicolumn{18}{r}{\textit{Continued on next page}} \\
\endfoot
\bottomrule
\endlastfoot

\multicolumn{18}{l}{\textit{Centralized}} \\
\midrule
YOLO26n & Full FT
& 0.777 & 0.964 & 0.165 & 0.635
& 0.752 & 0.937 & 0.199 & 0.630
& 0.679 & 0.863 & 0.301 & 0.614
& 0.714 & 0.899 & 0.240 & 0.617 \\
YOLO26s & Full FT
& 0.749 & 0.941 & 0.561 & 0.750
& 0.766 & 0.951 & 0.601 & 0.773
& 0.660 & 0.882 & 0.555 & 0.699
& 0.709 & 0.915 & 0.577 & 0.734 \\
YOLO26m & Full FT
& 0.622 & 0.921 & 0.549 & 0.697
& 0.710 & 0.903 & 0.586 & 0.733
& 0.550 & 0.856 & 0.521 & 0.642
& 0.620 & 0.879 & 0.552 & 0.684 \\
$DINOv2_{base}$ & Freeze
& 0.713 & 0.951 & 0.394 & 0.686
& 0.750 & 0.894 & 0.463 & 0.702
& 0.639 & 0.935 & 0.316 & 0.630
& 0.690 & 0.914 & 0.376 & 0.660 \\
$DINOv2_{base}$ & Full FT
& 0.660 & 0.960 & 0.281 & 0.633
& 0.806 & 0.903 & 0.291 & 0.667
& 0.555 & 0.950 & 0.598 & 0.701
& 0.657 & 0.926 & 0.392 & 0.658 \\
$DINOv2_{base}$ & Proposed
& 0.724 & 0.977 & 0.422 & 0.708
& 0.785 & 0.952 & 0.512 & 0.750
& 0.661 & 0.943 & 0.478 & 0.694
& 0.718 & 0.947 & 0.495 & 0.720 \\
$DINOv2_{fetal}$ & Freeze
& 0.782 & 0.977 & 0.385 & 0.715
& 0.769 & 0.894 & 0.494 & 0.719
& 0.731 & 0.962 & 0.496 & 0.730
& 0.749 & 0.927 & 0.495 & 0.724 \\
$DINOv2_{fetal}$ & Full FT
& 0.718 & 0.916 & 0.408 & 0.681
& 0.764 & 0.900 & 0.449 & 0.704
& 0.677 & 0.852 & 0.500 & 0.676
& 0.718 & 0.875 & 0.473 & 0.689 \\
$DINOv2_{fetal}$ & Proposed
& 0.752 & 0.983 & 0.608 & 0.781
& 0.797 & 0.956 & 0.633 & 0.795
& 0.645 & 0.920 & 0.537 & 0.701
& 0.713 & 0.938 & 0.581 & 0.744 \\
$DINOv3_{base}$ & Freeze
& 0.579 & 0.973 & 0.627 & 0.726
& 0.609 & 0.895 & 0.635 & 0.713
& 0.576 & 0.943 & 0.654 & 0.724
& 0.592 & 0.919 & 0.645 & 0.718 \\
$DINOv3_{base}$ & Full FT
& 0.732 & 0.955 & 0.636 & 0.774
& 0.776 & 0.884 & 0.704 & 0.788
& 0.677 & 0.939 & 0.552 & 0.723
& 0.723 & 0.911 & 0.619 & 0.751 \\
$DINOv3_{base}$ & Proposed
& 0.686 & 0.965 & 0.620 & 0.757
& 0.733 & 0.901 & 0.584 & 0.739
& 0.676 & 0.937 & 0.743 & 0.785
& 0.703 & 0.919 & 0.654 & 0.759 \\

\midrule
\multicolumn{18}{l}{\textit{Federated -- FedProx}} \\
\midrule
YOLO26n & Full FT
& 0.788 & 0.959 & 0.673 & 0.807
& 0.854 & 0.908 & 0.759 & 0.840
& 0.698 & 0.939 & 0.601 & 0.746
& 0.769 & 0.923 & 0.670 & 0.787 \\
YOLO26s & Full FT
& 0.834 & 0.975 & 0.518 & 0.775
& 0.859 & 0.941 & 0.563 & 0.787
& 0.801 & 0.901 & 0.493 & 0.732
& 0.829 & 0.921 & 0.525 & 0.758 \\
YOLO26m & Full FT
& 0.791 & 0.966 & 0.734 & 0.830
& 0.832 & 0.923 & 0.763 & 0.839
& 0.728 & 0.875 & 0.728 & 0.777
& 0.777 & 0.898 & 0.728 & 0.801 \\
SAM & Proposed
& 0.689 & 0.952 & 0.448 & 0.696
& 0.784 & 0.903 & 0.472 & 0.720
& 0.590 & 0.889 & 0.552 & 0.677
& 0.673 & 0.896 & 0.508 & 0.692 \\
$DINOv3_{base}$ & Freeze
& 0.788 & 0.974 & 0.711 & 0.824
& 0.818 & 0.908 & 0.662 & 0.796
& 0.666 & 0.962 & 0.676 & 0.768
& 0.734 & 0.934 & 0.669 & 0.779 \\
$DINOv3_{base}$ & Full FT
& 0.789 & 0.965 & 0.655 & 0.803
& 0.797 & 0.957 & 0.630 & 0.794
& 0.734 & 0.893 & 0.750 & 0.792
& 0.764 & 0.924 & 0.685 & 0.791 \\
$DINOv3_{base}$ & Proposed
& 0.772 & 0.976 & 0.702 & 0.816
& 0.767 & 0.949 & 0.713 & 0.809
& 0.709 & 0.920 & 0.684 & 0.771
& 0.737 & 0.934 & 0.698 & 0.790 \\
UltraFedFM & Proposed
& 0.695 & 0.974 & 0.061 & 0.576
& 0.748 & 0.943 & 0.132 & 0.607
& 0.603 & 0.901 & 0.191 & 0.565
& 0.668 & 0.921 & 0.156 & 0.582 \\
UltraFedFM & Freeze
& 0.679 & 0.968 & 0.186 & 0.611
& 0.751 & 0.846 & 0.319 & 0.639
& 0.599 & 0.983 & 0.471 & 0.684
& 0.666 & 0.910 & 0.381 & 0.652 \\
UltraFedFM & Full FT
& 0.675 & 0.969 & 0.243 & 0.629
& 0.858 & 0.845 & 0.359 & 0.687
& 0.590 & 0.977 & 0.449 & 0.672
& 0.699 & 0.906 & 0.399 & 0.668 \\
UltraSAM & Freeze
& 0.813 & 0.957 & 0.562 & 0.777
& 0.796 & 0.860 & 0.573 & 0.743
& 0.754 & 0.977 & 0.537 & 0.756
& 0.774 & 0.915 & 0.554 & 0.748 \\
UltraSAM & Full FT
& 0.796 & 0.979 & 0.614 & 0.796
& 0.808 & 0.973 & 0.542 & 0.775
& 0.683 & 0.901 & 0.610 & 0.732
& 0.741 & 0.936 & 0.574 & 0.750 \\
UltraSAM & Proposed
& 0.787 & 0.962 & 0.559 & 0.769
& 0.831 & 0.873 & 0.540 & 0.748
& 0.774 & 0.951 & 0.618 & 0.781
& 0.801 & 0.910 & 0.576 & 0.763 \\
$DINOv2_{fetal}$ & Freeze
& 0.670 & 0.961 & 0.472 & 0.701
& 0.836 & 0.850 & 0.474 & 0.720
& 0.584 & 0.954 & 0.559 & 0.699
& 0.687 & 0.899 & 0.513 & 0.700 \\
$DINOv2_{fetal}$ & Full FT
& 0.721 & 0.937 & 0.467 & 0.708
& 0.827 & 0.859 & 0.506 & 0.731
& 0.660 & 0.939 & 0.504 & 0.701
& 0.734 & 0.897 & 0.505 & 0.712 \\
$DINOv2_{fetal}$ & Proposed
& 0.741 & 0.965 & 0.727 & 0.811
& 0.827 & 0.949 & 0.654 & 0.810
& 0.624 & 0.882 & 0.757 & 0.754
& 0.711 & 0.914 & 0.702 & 0.776 \\
$DINOv2_{base}$ & Freeze
& 0.705 & 0.961 & 0.548 & 0.738
& 0.713 & 0.912 & 0.554 & 0.726
& 0.684 & 0.903 & 0.593 & 0.727
& 0.698 & 0.907 & 0.573 & 0.726 \\
$DINOv2_{base}$ & Full FT
& 0.742 & 0.969 & 0.601 & 0.771
& 0.735 & 0.935 & 0.622 & 0.764
& 0.702 & 0.926 & 0.641 & 0.756
& 0.718 & 0.930 & 0.631 & 0.760 \\
$DINOv2_{base}$ & Proposed
& 0.814 & 0.973 & 0.747 & 0.845
& 0.752 & 0.922 & 0.683 & 0.786
& 0.739 & 0.901 & 0.683 & 0.774
& 0.752 & 0.922 & 0.683 & 0.786 \\

\midrule
\multicolumn{18}{l}{\textit{Federated -- FedAvg}} \\
\midrule
YOLO26n & Full FT
& 0.790 & 0.972 & 0.594 & 0.785
& 0.877 & 0.930 & 0.637 & 0.815
& 0.678 & 0.970 & 0.566 & 0.738
& 0.765 & 0.949 & 0.599 & 0.771 \\
YOLO26s & Full FT
& 0.841 & 0.960 & 0.458 & 0.753
& 0.892 & 0.913 & 0.439 & 0.748
& 0.731 & 0.932 & 0.599 & 0.754
& 0.804 & 0.922 & 0.507 & 0.744 \\
YOLO26m & Full FT
& 0.802 & 0.977 & 0.416 & 0.732
& 0.869 & 0.967 & 0.457 & 0.764
& 0.727 & 0.890 & 0.515 & 0.710
& 0.791 & 0.927 & 0.485 & 0.734 \\
SAM & Proposed
& 0.673 & 0.955 & 0.427 & 0.685
& 0.778 & 0.899 & 0.454 & 0.710
& 0.567 & 0.894 & 0.538 & 0.666
& 0.656 & 0.896 & 0.492 & 0.681 \\
$DINOv3_{base}$ & Freeze
& 0.772 & 0.975 & 0.641 & 0.796
& 0.816 & 0.917 & 0.671 & 0.801
& 0.651 & 0.951 & 0.690 & 0.764
& 0.724 & 0.934 & 0.681 & 0.779 \\
$DINOv3_{base}$ & Full FT
& 0.774 & 0.977 & 0.696 & 0.816
& 0.803 & 0.874 & 0.674 & 0.784
& 0.697 & 0.981 & 0.676 & 0.785
& 0.746 & 0.925 & 0.675 & 0.782 \\
$DINOv3_{base}$ & Proposed
& 0.773 & 0.972 & 0.724 & 0.823
& 0.799 & 0.945 & 0.707 & 0.817
& 0.656 & 0.909 & 0.690 & 0.752
& 0.721 & 0.927 & 0.698 & 0.782 \\
UltraFedFM & Proposed
& 0.652 & 0.965 & 0.176 & 0.598
& 0.768 & 0.904 & 0.279 & 0.650
& 0.550 & 0.920 & 0.471 & 0.647
& 0.641 & 0.912 & 0.350 & 0.634 \\
UltraFedFM & Freeze
& 0.589 & 0.964 & 0.312 & 0.622
& 0.651 & 0.862 & 0.392 & 0.635
& 0.564 & 0.973 & 0.529 & 0.689
& 0.605 & 0.914 & 0.451 & 0.656 \\
UltraFedFM & Full FT
& 0.624 & 0.954 & 0.207 & 0.595
& 0.714 & 0.853 & 0.315 & 0.627
& 0.545 & 0.985 & 0.449 & 0.660
& 0.618 & 0.914 & 0.370 & 0.634 \\
UltraSAM & Freeze
& 0.805 & 0.969 & 0.491 & 0.755
& 0.834 & 0.857 & 0.405 & 0.698
& 0.752 & 0.977 & 0.721 & 0.817
& 0.791 & 0.913 & 0.518 & 0.741 \\
UltraSAM & Full FT
& 0.834 & 0.980 & 0.742 & 0.852
& 0.877 & 0.960 & 0.636 & 0.824
& 0.678 & 0.901 & 0.720 & 0.767
& 0.765 & 0.930 & 0.676 & 0.790 \\
UltraSAM & Proposed
& 0.732 & 0.973 & 0.483 & 0.729
& 0.760 & 0.893 & 0.534 & 0.729
& 0.657 & 0.958 & 0.591 & 0.735
& 0.705 & 0.924 & 0.561 & 0.730 \\
$DINOv2_{fetal}$ & Freeze
& 0.711 & 0.931 & 0.433 & 0.692
& 0.812 & 0.854 & 0.457 & 0.708
& 0.660 & 0.935 & 0.485 & 0.694
& 0.728 & 0.893 & 0.471 & 0.697 \\
$DINOv2_{fetal}$ & Full FT
& 0.688 & 0.970 & 0.507 & 0.722
& 0.830 & 0.857 & 0.632 & 0.773
& 0.617 & 0.973 & 0.441 & 0.677
& 0.708 & 0.912 & 0.520 & 0.713 \\
$DINOv2_{fetal}$ & Proposed
& 0.741 & 0.978 & 0.800 & 0.839
& 0.705 & 0.892 & 0.780 & 0.792
& 0.672 & 0.966 & 0.731 & 0.790
& 0.688 & 0.927 & 0.755 & 0.790 \\
$DINOv2_{base}$ & Freeze
& 0.681 & 0.957 & 0.507 & 0.715
& 0.692 & 0.905 & 0.516 & 0.704
& 0.657 & 0.892 & 0.564 & 0.704
& 0.674 & 0.898 & 0.539 & 0.704 \\
$DINOv2_{base}$ & Full FT
& 0.630 & 0.953 & 0.579 & 0.721
& 0.706 & 0.924 & 0.569 & 0.733
& 0.671 & 0.914 & 0.615 & 0.733
& 0.688 & 0.924 & 0.587 & 0.733 \\
$DINOv2_{base}$ & Proposed
& 0.806 & 0.977 & 0.788 & 0.857
& 0.761 & 0.957 & 0.846 & 0.855
& 0.746 & 0.890 & 0.646 & 0.761
& 0.753 & 0.922 & 0.732 & 0.803 \\
\end{longtable}
\end{landscape}

\bibliographystyle{elsarticle-num-names} 
\bibliography{cas-refs}

@article{hu2022lora,
  title={{LoRA}: Low-Rank Adaptation of Large Language Models},
  author={Hu, Edward J and Shen, Yelong and Wallis, Phillip and Allen-Zhu, Zeyuan and Li, Yuanzhi and Wang, Shean and Wang, Liang and Chen, Weizhu and others},
  journal={Iclr},
  volume={1},
  number={2},
  pages={3},
  year={2022}
}

@article{zhang2020value,
  title={The value of obstetric ultrasound in screening fetal nervous system malformation},
  author={Zhang, Na and Dong, Han and Wang, Ping and Wang, Zhihui and Wang, Yuchong and Guo, Zhiheng},
  journal={World Neurosurgery},
  volume={138},
  pages={645--653},
  year={2020},
  publisher={Elsevier}
}

@article{paladini2007sonographic,
  title={Sonographic examination of the fetal central nervous system: guidelines for performing the ‘basic examination’and the ‘fetal neurosonogram’},
  author={Paladini, Dario and Malinger, Gustavo and Monteagudo, Ana and Pilu, Gianluigi and Timor Tritsch, Ilan and Toi, Ants and others},
  journal={Ultrasound in Obstetrics \& Gynecology},
  volume={29},
  number={1},
  pages={109--116},
  year={2007}
}

@article{wang2024assessment,
  title={Assessment of the development of the central nervous system in fetuses with fetal growth restriction},
  author={Wang, Xiaohan and Wang, Chunli and Yang, Wenming and Yao, Qing and Zuo, Linhui},
  journal={Archives of Gynecology and Obstetrics},
  volume={310},
  number={6},
  pages={2963--2971},
  year={2024},
  publisher={Springer}
}

@article{fiorentino2023review,
  title={A review on deep-learning algorithms for fetal ultrasound-image analysis},
  author={Fiorentino, Maria Chiara and Villani, Francesca Pia and Di Cosmo, Mariachiara and Frontoni, Emanuele and Moccia, Sara},
  journal={Medical Image Analysis},
  volume={83},
  pages={102629},
  year={2023},
  publisher={Elsevier}
}

@article{pilu2016fetal,
  title={Fetal central nervous system anomalies},
  author={Pilu, Gianluigi and Alfirevic, Zarko},
  journal={Fetal Medicine},
  pages={81},
  year={2016},
  publisher={Cambridge University Press}
}

@article{lanzarone2025fetal,
  title={Fetal Corpus Callosum Anomalies: A Review of Underlying Genetic Disorders and Prenatal Testing Options},
  author={Lanzarone, Valeria and Eixarch, Elisenda and Borrell, Antoni},
  journal={Journal of Ultrasound in Medicine},
  volume={44},
  number={4},
  pages={637--652},
  year={2025},
  publisher={Wiley Online Library}
}

@article{marathu2024fetal,
  title={Fetal {MRI} analysis of corpus callosal abnormalities: Classification, and associated anomalies},
  author={Marathu, Kranthi K and Vahedifard, Farzan and Kocak, Mehmet and Liu, Xuchu and Adepoju, Jubril O and Bowker, Rakhee M and Supanich, Mark and Cosme-Cruz, Rosario M and Byrd, Sharon},
  journal={Diagnostics},
  volume={14},
  number={4},
  pages={430},
  year={2024},
  publisher={MDPI}
}

@article{burgos2020evaluation,
  title={Evaluation of deep convolutional neural networks for automatic classification of common maternal fetal ultrasound planes},
  author={Burgos-Artizzu, Xavier P and Coronado-Guti{\'e}rrez, David and Valenzuela-Alcaraz, Brenda and Bonet-Carne, Elisenda and Eixarch, Elisenda and Crispi, Fatima and Gratac{\'o}s, Eduard},
  journal={Scientific Reports},
  volume={10},
  number={1},
  pages={10200},
  year={2020},
  publisher={Nature Publishing Group UK London}
}

@article{santo2012counseling,
  title={Counseling in fetal medicine: agenesis of the corpus callosum},
  author={Santo, S and D'antonio, F and Homfray, T and Rich, P and Pilu, Gianluigi and Bhide, A and Thilaganathan, B and Papageorghiou, AT},
  journal={Ultrasound in Obstetrics \& Gynecology},
  volume={40},
  number={5},
  pages={513--521},
  year={2012},
  publisher={Wiley Online Library}
}

@article{torres2022review,
  title={A review of image processing methods for fetal head and brain analysis in ultrasound images},
  author={Torres, Helena R and Morais, Pedro and Oliveira, Bruno and Birdir, Cahit and R{\"u}diger, Mario and Fonseca, Jaime C and Vila{\c{c}}a, Jo{\~a}o L},
  journal={Computer Methods and Programs in Biomedicine},
  volume={215},
  pages={106629},
  year={2022},
  publisher={Elsevier}
}

@article{sendra2023generalisability,
  title={Generalisability of fetal ultrasound deep learning models to low-resource imaging settings in five African countries},
  author={Sendra-Balcells, Carla and Campello, V{\'\i}ctor M and Torrents-Barrena, Jordina and Ahmed, Yahya Ali and Elattar, Mustafa and Ohene-Botwe, Benard and Nyangulu, Pempho and Stones, William and Ammar, Mohammed and Benamer, Lamya Nawal and others},
  journal={Scientific Reports},
  volume={13},
  number={1},
  pages={2728},
  year={2023},
  publisher={Nature Publishing Group UK London}
}

@article{meng2020automatic,
  title={Automatic display of fetal brain planes and automatic measurements of fetal brain parameters by transabdominal three-dimensional ultrasound},
  author={Meng, Lu and Zhao, Dan and Yang, Zeyu and Wang, Bing},
  journal={Journal of Clinical Ultrasound},
  volume={48},
  number={2},
  pages={82--88},
  year={2020},
  publisher={Wiley Online Library}
}

@article{wang2023method,
  title={A method framework of automatic localization and quantitative segmentation for the cavum septum pellucidum complex and the cerebellar vermis in fetal brain ultrasound images},
  author={Wang, Qifeng and Pei, Jingzhu and Ouyang, Jing and Chen, Yanjie and Pu, Juncheng and Humayun, Ahsan and Zhao, Dan and Liu, Bin},
  journal={Quantitative Imaging in Medicine and Surgery},
  volume={13},
  number={9},
  pages={6059},
  year={2023}
}

@article{khan2025comprehensive,
  title={A comprehensive survey of foundation models in medicine},
  author={Khan, Wasif and Leem, Seowung and See, Kyle B and Wong, Joshua K and Zhang, Shaoting and Fang, Ruogu},
  journal={IEEE Reviews in Biomedical Engineering},
  year={2025},
  publisher={IEEE}
}

@inproceedings{zhang2025adapting,
  title={Adapting vision foundation models for real-time ultrasound image segmentation},
  author={Zhang, Xiaoran and Chen, Eric Z and Zhao, Lin and Chen, Xiao and Liu, Yikang and Maihe, Boris and Duncan, James S and Chen, Terrence and Sun, Shanhui},
  booktitle={International Conference on Medical Image Computing and Computer-Assisted Intervention},
  pages={24--34},
  year={2025},
  organization={Springer}
}

@article{jiao2024usfm,
  title={{Usfm}: A universal ultrasound foundation model generalized to tasks and organs towards label efficient image analysis},
  author={Jiao, Jing and Zhou, Jin and Li, Xiaokang and Xia, Menghua and Huang, Yi and Huang, Lihong and Wang, Na and Zhang, Xiaofan and Zhou, Shichong and Wang, Yuanyuan and others},
  journal={Medical Image Analysis},
  volume={96},
  pages={103202},
  year={2024},
  publisher={Elsevier}
}

@article{fiorentino2025adapt,
  title={Adapt or specialize? A comprehensive evaluation of adapted SAM versus task-specific CNNs for fetal abdominal segmentation},
  author={Fiorentino, Maria Chiara and Federici, Lorenzo and La Camera, Alessandro Pietro and Caiani, Enrico Gianluca},
  journal={Computer Methods and Programs in Biomedicine},
  pages={109178},
  year={2025},
  publisher={Elsevier}
}

@article{meyer2025ultrasam,
  title={Ultrasam: a foundation model for ultrasound using large open-access segmentation datasets},
  author={Meyer, Adrien and Murali, Aditya and Zarin, Farahdiba and Mutter, Didier and Padoy, Nicolas},
  journal={International Journal of Computer Assisted Radiology and Surgery},
  pages={1--10},
  year={2025},
  publisher={Springer}
}

@article{ma2025tinyusfm,
  title={{TinyUSFM}: Towards Compact and Efficient Ultrasound Foundation Models},
  author={Ma, Chen and Jiao, Jing and Liang, Shuyu and Fu, Junhu and Wang, Qin and Li, Zeju and Wang, Yuanyuan and Guo, Yi},
  journal={arXiv preprint arXiv:2510.19239},
  year={2025}
}

@article{egana2015neurosonographic,
  title={Neurosonographic assessment of the corpus callosum as imaging biomarker of abnormal neurodevelopment in late-onset fetal growth restriction},
  author={Egana-Ugrinovic, Gabriela and Savchev, Stefan and Baz{\'a}n-Arcos, Carolina and Puerto, Bienvenido and Gratacos, Eduard and Sanz-Cortes, Magdalena},
  journal={Fetal Diagnosis and Therapy},
  volume={37},
  number={4},
  pages={281--288},
  year={2015},
  publisher={S. Karger AG}
}

@article{huang2018learning,
  title={Learning to segment key clinical anatomical structures in fetal neurosonography informed by a region-based descriptor},
  author={Huang, Ruobing and Namburete, Ana and Noble, Alison},
  journal={Journal of Medical Imaging},
  volume={5},
  number={1},
  pages={014007--014007},
  year={2018},
  publisher={Society of Photo-Optical Instrumentation Engineers}
}

@article{wang2025fb,
  title={FB-ZWUNet: A deep learning network for corpus callosum segmentation in fetal brain ultrasound images for prenatal diagnostics},
  author={Wang, Qifeng and Zhao, Dan and Ma, Hao and Liu, Bin},
  journal={Biomedical Signal Processing and Control},
  volume={104},
  pages={107499},
  year={2025},
  publisher={Elsevier}
}

@article{li14deep,
  title={Deep Learning-Based Automated Detection of Fetal Corpus Callosum Abnormalities in Prenatal Ultrasound},
  author={Li, Min and Liu, Shizhen and Zhang, Zhonglu and Li, Qiang and Xu, Xuan},
  journal={Frontiers in Pediatrics},
  volume={14},
  pages={1774586},
  publisher={Frontiers}, 
  year={2026}
}

@inproceedings{ambsdorf2025general,
  title={General methods make great domain-specific foundation models: A case-study on fetal ultrasound},
  author={Ambsdorf, Jakob and Munk, Asbj{\o}rn and Llambias, Sebastian and Christensen, Anders N and Mikolaj, Kamil and Balestriero, Randall and Tolsgaard, Martin G and Feragen, Aasa and Nielsen, Mads},
  booktitle={International Conference on Medical Image Computing and Computer-Assisted Intervention},
  pages={271--281},
  year={2025},
  organization={Springer}
}

@article{zou2023object,
  title={Object detection in 20 years: A survey},
  author={Zou, Zhengxia and Chen, Keyan and Shi, Zhenwei and Guo, Yuhong and Ye, Jieping},
  journal={Proceedings of the IEEE},
  volume={111},
  number={3},
  pages={257--276},
  year={2023},
  publisher={IEEE}
}

@article{oquab2023dinov2,
  title={{DINOv2}: Learning robust visual features without supervision},
  author={Oquab, Maxime and Darcet, Timoth{\'e}e and Moutakanni, Th{\'e}o and Vo, Huy and Szafraniec, Marc and Khalidov, Vasil and Fernandez, Pierre and Haziza, Daniel and Massa, Francisco and El-Nouby, Alaaeldin and others},
  journal={arXiv preprint arXiv:2304.07193},
  year={2023}
}

@inproceedings{sohan2024review,
  title={A review on {YOLOv8} and its advancements},
  author={Sohan, Mupparaju and Sai Ram, Thotakura and Rami Reddy, Ch Venkata},
  booktitle={International Conference on Data Intelligence and Cognitive Informatics},
  pages={529--545},
  year={2024},
  organization={Springer}
}

@software{Jocher_Ultralytics_YOLO_2023,
  author  = {Jocher, Glenn and Chaurasia, Ayush and Qiu, Jing},
  title   = {{Ultralytics YOLO}},
  version = {8.0.0},
  year    = {2023},
  url     = {https://github.com/ultralytics/ultralytics},
  license = {AGPL-3.0}
}

@article{conti2025challenging,
  title={Challenging {DINOv3} Foundation Model under Low Inter-Class Variability: A Case Study on Fetal Brain Ultrasound},
  author={Conti, Edoardo and Rosati, Riccardo and Federici, Lorenzo and Mancini, Adriano and Fiorentin, Maria Chiara},
  journal={arXiv preprint arXiv:2511.01915},
  year={2025}
}

@article{zhou2025segment,
  title={Segment anything model for fetal head-pubic symphysis segmentation in intrapartum ultrasound image analysis},
  author={Zhou, Zihao and Lu, Yaosheng and Bai, Jieyun and Campello, Victor M and Feng, Fan and Lekadir, Karim},
  journal={Expert Systems with Applications},
  volume={263},
  pages={125699},
  year={2025},
  publisher={Elsevier}
}

@inproceedings{le2026viscera,
  title={{VISCERA-SAM}: Adapting Segment Anything for Multi-Visceral Fetal Abdominal Ultrasound Segmentation},
  author={Le, Minh HN and Pham, Khoa D and Vinh, Tuan and Nguyen, Thanh-Huy and Huynh, Han H and Le, Khanh TQ and Vu, Anh Mai and Luong, Ha NT and Bagci, Ulas and Xu, Min and others},
  booktitle={Medical Imaging with Deep Learning},
  year={2026}
}

@article{jiang2025ultrasam,
  title={{UltraSAM}: A foundational medical ultrasound segmentation model with limited training data},
  author={Jiang, Tao and Li, Yifang and Xing, Wenyu and Cao, Ran and Yu, Ming and Zhu, Yunkai and Chen, Yaqing and Li, Boyi and Ta, Dean},
  journal={Expert Systems with Applications},
  pages={130223},
  year={2025},
  publisher={Elsevier}
}

@article{jiang2025pretraining,
  title={From pretraining to privacy: federated ultrasound foundation model with self-supervised learning},
  author={Jiang, Yuncheng and Feng, Chun-Mei and Ren, Jinke and Wei, Jun and Zhang, Zixun and Hu, Yiwen and Liu, Yunbi and Sun, Rui and Tang, Xuemei and Du, Juan and others},
  journal={npj Digital Medicine},
  volume={8},
  number={1},
  pages={714},
  year={2025},
  publisher={Nature Publishing Group UK London}
}

@article{sapkota2025yolo26,
  title={YOLO26: key architectural enhancements and performance benchmarking for real-time object detection},
  author={Sapkota, Ranjan and Cheppally, Rahul Harsha and Sharda, Ajay and Karkee, Manoj},
  journal={arXiv preprint arXiv:2509.25164},
  year={2025}
}

@article{cha2022altered,
  title={Altered microstructure of the splenium of corpus callosum is associated with neurodevelopmental impairment in preterm infants with necrotizing enterocolitis},
  author={Cha, Jong Ho and Lim, Jung-Sun and Jang, Yong Hun and Hwang, Jae Kyoon and Na, Jae Yoon and Lee, Jong-Min and Lee, Hyun Ju and Ahn, Ja-Hye},
  journal={Italian Journal of Pediatrics},
  volume={48},
  number={1},
  pages={6},
  year={2022},
  publisher={Springer}
}

@article{lubian2024corpus,
  title={Corpus callosum long-term biometry in very preterm children related to cognitive and motor outcomes},
  author={Lubi{\'a}n-Guti{\'e}rrez, Manuel and Benavente-Fern{\'a}ndez, Isabel and Mar{\'\i}n-Almagro, Yolanda and Jim{\'e}nez-Luque, Natalia and Zuazo-Ojeda, Amaya and S{\'a}nchez-Sandoval, Yolanda and Lubi{\'a}n-L{\'o}pez, Sim{\'o}n P},
  journal={Pediatric Research},
  volume={96},
  number={2},
  pages={409--417},
  year={2024},
  publisher={Nature Publishing Group US New York}
}

@inproceedings{mcmahan2017communication,
  title={Communication-efficient learning of deep networks from decentralized data},
  author={McMahan, Brendan and Moore, Eider and Ramage, Daniel and Hampson, Seth and y Arcas, Blaise Aguera},
  booktitle={Artificial Intelligence and Statistics},
  pages={1273--1282},
  year={2017},
  organization={Pmlr}
}

@article{li2020federated,
  title={Federated optimization in heterogeneous networks},
  author={Li, Tian and Sahu, Anit Kumar and Zaheer, Manzil and Sanjabi, Maziar and Talwalkar, Ameet and Smith, Virginia},
  journal={Proceedings of Machine learning and Systems},
  volume={2},
  pages={429--450},
  year={2020}
}

@article{judi2026fb,
  title={{FB-UNet++}: federated biometric UNet++ model for segmentation and classification network of fetal anomaly detection in prenatal care},
  author={Judi, A Alfina and Suresh, P and Raj, T Ajith Bosco},
  journal={Biomedical Signal Processing and Control},
  volume={112},
  pages={108706},
  year={2026},
  publisher={Elsevier}
}

@article{sheller2020federated,
  title={Federated learning in medicine: facilitating multi-institutional collaborations without sharing patient data},
  author={Sheller, Micah J. and Edwards, Brandon and Reina, G. Anthony and Martin, Jason and Pati, Sarthak and Kotrotsou, Aikaterini and Milchenko, Mikhail and Xu, Weilin and Marcus, Daniel and Colen, Rivka R. and Bakas, Spyridon},
  journal={Scientific Reports},
  volume={10},
  number={1},
  pages={12598},
  year={2020},
  publisher={Nature Publishing Group},
  doi={10.1038/s41598-020-69250-1}
}

@article{bian2025survey,
  title={A survey on parameter-efficient fine-tuning for foundation models in federated learning},
  author={Bian, Jieming and Peng, Yuanzhe and Wang, Lei and Huang, Yin and Xu, Jie},
  journal={arXiv preprint arXiv:2504.21099},
  year={2025}
}

@article{fiorentino2025uncovering,
  title={Uncovering ethical biases in publicly available fetal ultrasound datasets},
  author={Fiorentino, Maria Chiara and Moccia, Sara and Cosmo, Mariachiara Di and Frontoni, Emanuele and Giovanola, Benedetta and Tiribelli, Simona},
  journal={npj Digital Medicine},
  volume={8},
  number={1},
  pages={355},
  year={2025},
  publisher={Nature Publishing Group UK London}
}

@inproceedings{karimireddy2020scaffold,
  title={Scaffold: Stochastic controlled averaging for federated learning},
  author={Karimireddy, Sai Praneeth and Kale, Satyen and Mohri, Mehryar and Reddi, Sashank and Stich, Sebastian and Suresh, Ananda Theertha},
  booktitle={International Conference on Machine Learning},
  pages={5132--5143},
  year={2020},
  organization={PMLR}
}

@article{han2026federated,
  title={Federated learning for prenatal detection of interrupted aortic arch using fetal ultrasound imaging},
  author={Han, Jiancheng and Wang, Heqing and Feng, Yifan and Yang, Qi and Li, Jingtan and Zhang, Haojie and He, Yihua and Liu, Jiang and Nakamura, Toru and Cao, Yang and others},
  journal={Biomedical Signal Processing and Control},
  volume={119},
  pages={109795},
  year={2026},
  publisher={Elsevier}
}

@article{fiorentino2025contrastive,
  title={Contrastive prototype federated learning against noisy labels in fetal standard plane detection},
  author={Fiorentino, Maria Chiara and Migliorelli, Giovanna and Villani, Francesca Pia and Frontoni, Emanuele and Moccia, Sara},
  journal={International Journal of Computer Assisted Radiology and Surgery},
  volume={20},
  number={7},
  pages={1431--1439},
  year={2025},
  publisher={Springer}
}

@misc{eu2024aiact,
  title={{Regulation (EU) 2024/1689 of the European Parliament and of the Council of 13 June 2024 laying down harmonised rules on artificial intelligence (Artificial Intelligence Act)}},
  author={{European Parliament and Council of the European Union}},
  year={2024},
  howpublished={Official Journal of the European Union},
  url={https://eur-lex.europa.eu/eli/reg/2024/1689/oj},
  note={Accessed: 2026-05-13}
}
\end{document}